%% file: paper.tex
\documentclass[]{bytedance_seed}
\usepackage{helvet}
\usepackage{amsmath} 
\usepackage{natbib}
\usepackage{graphicx}
\usepackage{subcaption}

\usepackage[toc,page,header]{appendix}
\usepackage[utf8]{inputenc} % allow utf-8 input
\usepackage[T1]{fontenc}    % use 8-bit T1 fonts
\usepackage{hyperref}       % hyperlinks
\usepackage{url}            % simple URL typesetting
\usepackage{booktabs}       % professional-quality tables
\usepackage{lmodern}        % scalable Latin Modern fonts
\usepackage{amsfonts}       % blackboard math symbols
\usepackage{nicefrac}       % compact symbols for 1/2, etc.
\usepackage{microtype}      % microtypography
\usepackage{wrapfig} 

\usepackage{amssymb}  % 用于特殊符号如♠♣等
\usepackage{fontawesome}  % 用于邮箱图标 \faEnvelope
\usepackage{url}  % 用于 \url 命令
\usepackage{minitoc}

\usepackage{booktabs}
\usepackage{array}
\usepackage{etoolbox}

\definecolor{lightblue}{RGB}{200, 230, 255}  
\definecolor{headerblue}{RGB}{150, 200, 255}

\title{First Return, Entropy-Eliciting Explore}
\author[1,2,*]{Tianyu Zheng}
\author[1,2,*]{Tianshun Xing}
\author[1,*]{Qingshui Gu}
\author[2,*]{Taoran Liang}
\author[1,3]{Xingwei Qu}
\author[1]{Xin Zhou}
\author[2,3]{Yizhi Li}
\author[1]{Zhoufutu Wen}
\author[3]{Chenghua Lin}
\author[1]{Wenhao Huang}
\author[1,\dagger]{Qian Liu}
\author[1,2,\dagger]{Ge Zhang}
\author[1]{Zejun Ma}

\affiliation[1]{ByteDance}
\affiliation[2]{M-A-P}
\affiliation[3]{The University of Manchester}

\contribution[*]{Equal contribution}
\contribution[\dagger]{Corresponding authors}

\abstract{
\begin{abstract}

Reinforcement Learning from Verifiable Rewards (RLVR) improves the reasoning abilities of Large Language Models (LLMs) but it struggles with unstable exploration. We propose \textbf{FR3E} (First Return, Entropy-Eliciting Explore), a structured exploration framework that identifies high-uncertainty decision points in reasoning trajectories and performs targeted rollouts to construct semantically grounded intermediate feedback. Our method provides targeted guidance without relying on dense supervision. Empirical results on mathematical reasoning benchmarks(AIME24) show that FR3E promotes more stable training, produces longer and more coherent responses, and increases the proportion of fully correct trajectories. These results highlight the framework's effectiveness in improving LLM reasoning through more robust and structured exploration.
\end{abstract}
}

\date{\today}
\correspondence{Qian Liu at \email{qian.liu@bytedance.com}, Ge Zhang at \email{zhangge.eli@bytedance.com}}

\checkdata[Project Page]{\url{https://huggingface.co/FR3E-Bytedance}}

\begin{document}
\maketitle

%不需要目录就注释掉 注意目录不要和第一页放在一块 要有\newpage
%\newpage
%\tableofcontents
%\newpage

\input{sections/introduction}
\input{sections/relatedwork}
\input{sections/preliminary}
\input{sections/method}
\input{sections/experiments}
\input{sections/conclusion}

\clearpage

\bibliographystyle{plainnat}
\bibliography{main}

\clearpage

\beginappendix

\input{sections/appendix}

\end{document}

%% file: sections/introduction.tex
\section{Introduction}

Reinforcement Learning (RL) significantly enhances the reasoning capabilities~\cite{brown2020language, zhou2022least, wei2022chain, ouyang2022training, wu2024comparative, forootani2025survey, huan2025doesmathreasoningimprove} of LLMs, particularly in complex tasks like mathematical problem-solving and code generation using RLVR~\citep{shao2024deepseekmathpushinglimitsmathematical, yu2025dapoopensourcellmreinforcement, yue2025vapoefficientreliablereinforcement,guo2025deepseek}. A central challenge in these RLVR methodologies is the granular assignment of credit to intermediate steps within a reasoning trajectory. Many approaches, such as Group Relative Policy Optimization (GRPO)~\citep{shao2024deepseekmathpushinglimitsmathematical}, use the final outcome reward to estimate the value of all intermediate actions. This method, while simple, incorrectly averages the contribution of each step, failing to distinguish pivotal reasoning from inconsequential actions. Such uniform credit assignment limits the model's learning potential and deviates from human intuition where steps have varying importance. For tasks with sparse and delayed rewards, this leads to imprecise credit assignment, hindering the model's ability to identify crucial actions and potentially causing issues like ``overthinking''~\citep{chen2025think23overthinkingo1like}.

Existing strategies for estimating intermediate state or action values have notable limitations. Value model-based approaches like Proximal Policy Optimization (PPO)~\citep{schulman2017proximalpolicyoptimizationalgorithms} and Value-model Augmented Policy Optimization (VAPO)~\citep{yue2025vapoefficientreliablereinforcement} employ a critic model. However, training a stable and accurate critic for the vast state space of LLMs is difficult, often leading to instability, bias, and significant computational overhead. Even with techniques like Generalized Advantage Estimation (GAE)~\citep{schulman2018highdimensionalcontinuouscontrolusing}, reward propagation to distant tokens remains problematic. While GRPO avoids a critic, it primarily relies on trajectory-level rewards, not fully solving granular credit assignment. Other methods use sampling or heuristic-based intermediate reward estimation. For instance, VinePPO~\citep{kazemnejad2024vineppounlockingrlpotential} segments trajectories heuristically and uses Monte Carlo (MC) sampling for step-level advantages. Process reward models (PRMs)~\citep{luo2024improvemathematicalreasoninglanguage,lightman2023letsverifystepstep}, as in S-GRPO~\citep{dai2025sgrpoearlyexitreinforcement}, and PRIME~\citep{cui2025processreinforcementimplicitrewards}, aim to provide intermediate feedback but face challenges with heuristic quality, sampling variance, labeling costs for explicit PRMs, or the granularity of implicitly derived rewards. Fixed reward schemes may also lack adaptability. Current methods thus struggle with complexity, instability, cost, or reliance on brittle heuristics.

In this work, we introduce \textbf{First Return, Entropy-Eliciting Explore} (FR3E), a structured
exploration framework that identifies high-uncertainty decision points in reasoning trajectories and performs targeted rollouts to construct semantically grounded intermediate feedback. Unlike traditional methods that perform full rollouts starting directly from prompts, FR3E localizes high-entropy tokens along correctly completed reasoning trajectories as critical decision points. These tokens serve as anchors from which targeted rollouts are initiated, enabling structured exploration around uncertain yet pivotal steps in the reasoning process.

By performing partial rollouts from these key decision points, FR3E synthesizes localized feedback signals that would otherwise be unavailable in standard autoregressive generation. This process does not depend on existing reward signals or attempt to perform credit assignment over long sequences; instead, it actively constructs new sources of evaluation through controlled exploration. These synthesized signals are semantically grounded and particularly useful for guiding early-stage reasoning, where feedback is often most ambiguous.

The design of FR3E builds upon the principles of ``First Return, Then Explore'' (FRTE)~\citep{Ecoffet_2021}, adapted to the sequential generation behavior of LLMs. Compared to conventional reinforcement learning approaches that often treat all generation steps uniformly, FR3E leverages entropy profiles to identify critical reasoning junctures. It then performs targeted rollouts from these states, yielding structured exploration and more meaningful feedback, all without requiring detailed reward labels for every step.

We summarize our key contributions as follows:
\begin{enumerate}
\item \textbf{A Reliable Exploration Framework for Trajectory-Level Reward Shaping}: We introduce FR3E, a novel reinforcement learning algorithm that improves reward shaping at the trajectory level by emphasizing reliable exploration paths. 

\item \textbf{Improved Training Stability and Reasoning Capability}: FR3E maintains entropy at a stable and gradually increasing level during training, preventing early collapse and enabling the model to generate longer and more reliable reasoning chains. This addresses common failures in standard RL, particularly in specialized models such as Qwen-Math-7B.

\item \textbf{More Robust Positive Reward Signals}: The core mechanism of FR3E generates more positive reward signals by encouraging structured exploration around high-entropy states. Empirical analysis over multiple rollouts per prompt shows that a higher proportion of generated trajectories are fully correct (``All-Right''), while fewer are completely incorrect (``All-Wrong'').
\end{enumerate}

By combining uncertainty-driven exploration with sampled intermediate feedback, FR3E supports a more stable and data-efficient RL training process. Its value-model-free design encourages meaningful trajectory expansion without relying on brittle reward shaping or complex critics.

%\subsection{Hello World}

%% file: sections/relatedwork.tex
\section{Related Work}

\textbf{Reinforcement Learning: Challenges and Techniques}

RL faces key challenges when applied to sparse-reward or long-horizon tasks. Simple exploration strategies like $\epsilon$-greedy and Upper Confidence Bound (UCB)~\citep{auer2002using,auer2002finite} are often ineffective in large state spaces. Structured methods such as Go-Explore~\citep{ecoffet2019go, Ecoffet_2021} improve performance by archiving and revisiting promising states---a principle that inspires our work; its extension, Intelligent Go-Explore (IGE)~\citep{lu2025intelligentgoexplorestandingshoulders}, uses foundation models to guide exploration. To handle temporal abstraction, Hierarchical RL (HRL) decomposes tasks into subgoals via “options”~\citep{sutton1999between}. This idea underpins methods that assign intermediate rewards~\citep{setlur2024rewardingprogressscalingautomated}, such as Process Reward Models (PRMs)~\citep{luo2024improvemathematicalreasoninglanguage,lightman2023letsverifystepstep}, to guide learning and mitigate the credit assignment problem (CAP)~\citep{pignatelli2023survey}. CALM~\citep{luo2024improvemathematicalreasoninglanguage} further automates this by using LLMs to identify subgoals and assign auxiliary rewards, improving credit attribution and training efficiency.

\textbf{Reinforcement Learning for Large Language Models}

RL enables LLMs to go beyond imitation learning and optimize generation based on task-specific rewards~\citep{ouyang2022training}. While PPO~\citep{schulman2017proximalpolicyoptimizationalgorithms} is widely used, its reliance on a value network is computationally costly and notoriously unstable for the vast, discrete action space of language generation. This instability motivates the development of value-model-free alternatives like GRPO~\citep{shao2024deepseekmathpushinglimitsmathematical} and RLOO~\citep{ahmadian2024basicsrevisitingreinforcestyle}, which uses trajectory-level comparisons, and S-GRPO~\citep{dai2025sgrpoearlyexitreinforcement}, which encourages conciseness with reward decay. Concurrently, other works aim to improve credit assignment granularity. VinePPO~\citep{kazemnejad2024vineppounlockingrlpotential}, for instance, uses Monte Carlo rollouts from heuristically chosen intermediate states, while PRIME~\citep{cui2025prime} derives implicit process rewards from final outcomes to reduce reliance on costly labels. While these methods advance granular credit assignment, adapting structured exploration strategies like Go-Explore to the autoregressive nature of LLMs remains an active research area. 
% Our work contributes to this direction by proposing a framework that explicitly identifies high-uncertainty decision points for targeted exploration, integrating this strategy within a stable, value-model-free algorithm.

%\subsection{Hello World}

%% file: sections/preliminary.tex
\section{Preliminary}

\label{sec:preliminary}
\subsection{MDP Setup for Autoregressive Generation}
Many language-based reasoning tasks can be naturally formulated as sequential decision-making problems, modeled as Markov Decision Processes (MDPs). An MDP is defined by the tuple \((\mathcal{S}, \mathcal{A}, P, r, \gamma)\), where \(\mathcal{S}\) is the state space, \(\mathcal{A}\) the action space, \(P\) the transition dynamics, \(r\) the reward function, and \(\gamma \in (0,1)\) the discount factor~\citep{sutton1998reinforcement}. The objective is to learn a policy \(\pi\) that maximizes the expected return:
\begin{equation}
\mathbb{E}_\pi \left[\sum_t \gamma^t r(s_t, a_t)\right]
\end{equation}

In the context of large language models (LLMs), the environment corresponds to the autoregressive generation process~\citep{81a0972a0b704d918ec2ce3696c5b871,ouyang2022training}. Specifically:

\textbf{States (\(\mathcal{S}\)):} Each state \(s_t = (x_0, \dots, x_m, y_0, \dots, y_t)\) includes the input context \(x_{0:m}\) and the generated tokens \(y_{0:t}\).

\textbf{Actions (\(\mathcal{A}\)):} An action \(a = y_{t+1} \in \mathcal{V}\) corresponds to selecting the next token from the vocabulary \(\mathcal{V}\).

\textbf{Dynamics (\(P\)):} The environment transitions deterministically to \(s_{t+1} = (x_0, \dots, x_m, y_0, \dots, y_{t+1})\).

\textbf{Reward (\(r\)):} Rewards are typically sparse, provided only at the end of a complete rollout, limiting intermediate feedback and posing challenges for effective exploration~\citep{schulman2017proximalpolicyoptimizationalgorithms}.

\subsection{Rejection Sampling}

In the stage of advantage estimating, when a prompt consistently yields trajectories with identical rewards, either all correct (reward = 1) or all incorrect (reward = 0), the corresponding advantage estimates become zero. This leads to vanishing policy gradients and results in reduced sample efficiency. As training progresses, more prompts fall into this degenerate category, causing the effective batch size to shrink and increasing the variance of gradient estimates.

To mitigate this issue during our \textit{entropy finding} phase—which aims to identify uncertain reasoning positions for structured exploration—we employ \textit{rejection sampling}~\citep{yu2025dapoopensourcellmreinforcement,zhang2025srpocrossdomainimplementationlargescale}. Specifically, for each prompt, we perform multiple rollouts and compute their corresponding binary rewards. If all rollouts yield the same reward (i.e., either all 0s or all 1s), the prompt is considered degenerate and is rejected from the current batch. This ensures that the entropy-finding process operates on diverse and informative reasoning trajectories, rather than being misled by degenerate cases with no variation in reward.

\subsection{Clip-Higher}
Standard PPO-based reinforcement learning constrains policy updates using symmetric clipping bounds, typically limiting the probability ratio $r_{i,t}(\theta)$ within $[0.8, 1.2]$ to prevent unstable updates. However, this symmetric structure also suppresses the probability growth of low-confidence tokens, which hinders exploration and can lead to \textit{entropy collapse}---a state where the policy becomes overly deterministic and loses action diversity.

To address this issue and explicitly encourage exploration, we adopt the \textit{Clip-Higher} mechanism~\citep{yue2025vapoefficientreliablereinforcement}, which replaces symmetric clipping with asymmetric bounds that permit more substantial increases in the probability of underexplored but potentially valuable reasoning paths, while still limiting overly aggressive decreases. This asymmetry encourages the policy to better explore less probable trajectories without sacrificing training stability. Empirical validation in our setting shows that this design improves both exploration efficiency and learning robustness. We use $\epsilon_{\text{low}} = 0.22$ and $\epsilon_{\text{high}} = 0.28$, yielding the objective:
\begin{equation}
\mathcal{L}_{\text{PPO}}(\theta) =
-\frac{1}{\sum_{i=1}^G |o_i|}
\sum_{i=1}^G \sum_{t=1}^{|o_i|}
\min\left(
r_{i,t}(\theta) \hat{A}_{i,t},~
\text{clip}(r_{i,t}(\theta), 1 - \epsilon_{\text{low}}, 1 + \epsilon_{\text{high}}) \hat{A}_{i,t}
\right)
\end{equation}
This asymmetrically clipped objective plays a key role in preserving policy diversity and enhancing long-horizon reasoning through more effective exploration.

% \subsection{Structured Exploration}

% The principle of structured exploration, which involves systematically revisiting and expanding upon promising partial solutions, has demonstrated considerable efficacy in reinforcement learning, exemplified by seminal approaches like Go-Explore~\citep{Ecoffet_2021}. This paradigm is particularly pertinent to LLMs operating in complex reasoning tasks characterized by sparse and delayed rewards. In such scenarios, conventional random exploration often proves inefficient, failing to identify or adequately exploit critical intermediate reasoning steps that are essential for reaching optimal solutions.

% To address this limitation, we investigate a targeted, two-stage exploration strategy. The initial stage focuses on identifying high-uncertainty decision points within nascent reasoning trajectories. These points, often corresponding to local minima in token-level log-probabilities, signify potential junctures or critical forks in the reasoning process and are designated as intermediate reasoning states. In the subsequent stage, these identified intermediate states are systematically revisited, and controlled partial rollouts are initiated from them to expand upon the existing reasoning paths. This structure provide a principled framework for navigating the reasoning space of LLMs, enabling the model to build upon promising partial solutions while efficiently discovering new, high-reward trajectories.

\subsection{Structured Exploration}
Our methodology is inspired by the principles of structured exploration, particularly the ``First Return, Then Explor'' paradigm popularized by Go-Explore~\citep{Ecoffet_2021}. This approach has demonstrated considerable efficacy in tasks with sparse rewards by first focusing on finding a path to a novel state (the ``first return'') and only then initiating deeper exploration from that promising location. For complex reasoning tasks in LLMs, where random exploration is inefficient, this paradigm offers a more principled way to navigate the vast generation space.

We adapt this ``First Return, Then Explore'' principle for the autoregressive generation process of LLMs. The \textbf{``First Return''} phase involves generating initial, promising trajectories. Instead of exploring randomly, our \textbf{``Entropy-Eliciting Explore''} phase then analyzes these trajectories to pinpoint states of high model uncertainty. These states, often identified by local minima in token-level log-probabilities (i.e., high local entropy), represent critical junctures where the model is least confident. By systematically initiating partial rollouts from these high-entropy states, we can efficiently discover valuable alternative reasoning paths and generate dense, informative feedback, enabling the model to build upon promising solutions in a targeted manner.

%% file: sections/method.tex
\section{FR3E: First Return, Entropy-Eliciting Explore}
\label{sec:method}
Effective exploration in LLM reinforcement learning faces two primary obstacles: the loss of promising search trajectories and the inability to return to productive reasoning paths. Our approach, \textbf{FR3E}, addresses these challenges through a structured framework that explicitly separates the discovery of high-potential reasoning points from their systematic exploration.

The core idea is to decompose the reinforcement learning process into two complementary phases:
\begin{itemize}
    \item \textbf{First Return}: Identify key positions along base trajectories where model uncertainty is highest — potential forks in the reasoning space.
    \item \textbf{Entropy-Eliciting Explore}: From these identified states, perform diverse rollouts to sample alternative solution paths while maintaining semantic coherence.
\end{itemize}

This structured paradigm enhances traditional exploration-exploitation trade-offs by incorporating entropy-based signals directly into both path selection and policy update mechanisms.

\begin{figure}[ht]
    \centering
    \includegraphics[width=0.99\linewidth]{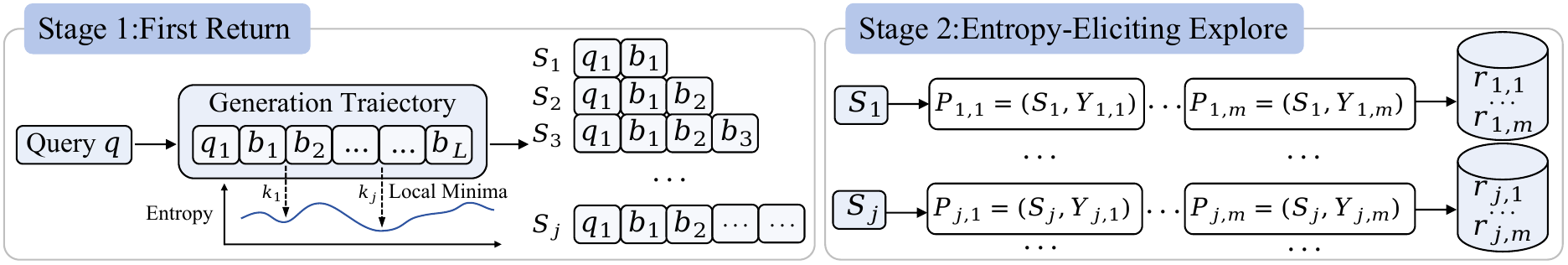}
    \caption{
Overview of the FR3E framework. \textbf{Stage 1: First Return} begins with base trajectory generation from query $q$, followed by token-wise entropy computation to identify high-uncertainty positions. These positions serve as segmentation points for constructing intermediate semantic states $S_j$. \textbf{Stage 2: Entropy-Eliciting Explore} launches multiple rollouts from each state $S_j$, evaluates the reward for each, and computes empirical values $V(S_j)$ to guide adaptive policy updates. This two-stage design encourages diverse yet structured exploration based on model uncertainty signals.
}

    \label{fig:main}
\end{figure}

\subsection{First Return: Block Definition via Uncertainty Signals}

\subsubsection{Base Trajectory Generation}

Given a training query $ q $, we first generate a base reasoning trajectory using the policy $ \pi_\theta $:
\begin{equation}
P_{\text{base}} = (q, t_1, t_2, \dots, t_L)
\end{equation}
where each $ t_i $ denotes a generated token in the response sequence. This trajectory represents the model's step-by-step reasoning process from the initial query to the final output.

\subsubsection{Entropy Computation}

To identify positions in the trajectory that exhibit high uncertainty — and thus are suitable for exploration — we compute the token-wise entropy at each position $ k $. Specifically, let:
\begin{equation}
\pi_\theta(v \mid q, t_{<k})
\end{equation}
denote the softmax-normalized probability distribution over the vocabulary at step $ k $, conditioned on the query $ q $ and previously generated tokens $ t_{<k} $.

We define the entropy of the policy at position $ k $ as:
\begin{equation}
H_k = -\sum_{v \in \mathcal{V}} \pi_\theta(v \mid q, t_{<k}) \log \pi_\theta(v \mid q, t_{<k})
\end{equation}
where $ \mathcal{V} $ is the model's vocabulary set.

Higher values of $ H_k $ indicate greater uncertainty in the model’s decision-making process at that position. These high-entropy positions represent promising candidates for initiating structured exploration.

\subsubsection{Entropy-Sensitive Position Selection}

Given the full base trajectory $ P_{\text{base}} $, we identify positions with maximal uncertainty by selecting the top-$ K $ tokens with highest entropy values. Specifically, we extract a set of entropy-sensitive positions:
\begin{equation}
\mathcal{K} = \{k_1, \dots, k_K\}
\end{equation}
which correspond to the most uncertain decision points along the reasoning path.

\textbf{Top-$ K $ Entropy Selection}: 
Select the indices corresponding to the $ K $ largest entropy values across the entire trajectory~\citep{wang20258020rulehighentropyminority}:
\begin{equation}
\mathcal{K} = \operatorname{TopK}\left( \{ H_k \}_{k=1}^L \right)
\end{equation}

This strategy ensures that we focus on the most uncertain reasoning steps globally, enabling us to identify natural forks in the model's reasoning space for structured exploration.

\newpage

\subsubsection{Semantic Block \& State Construction}

\begin{wrapfigure}{r}{0.5\textwidth}  % 'r' 表示右侧，宽度为50%
    \centering
    \vspace{-10pt}  % 可选，微调垂直位置
    \includegraphics[width=0.49\textwidth]{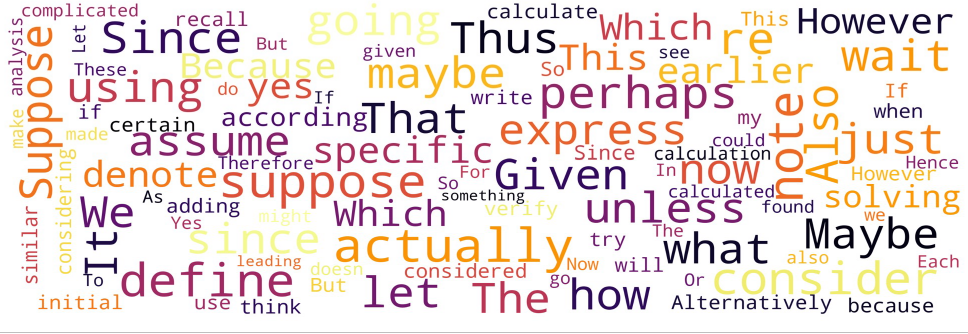}
    \caption{Frequent tokens with the highest average entropy}
    \label{fig:high_entropy}
    \vspace{-10pt}  % 可选
\end{wrapfigure}
As shown in Figure~\ref{fig:high_entropy}, frequent tokens with high average entropy can be identified and used as key segmentation points. This figure is adapted from prior work~\citep{wang20258020rulehighentropyminority}, which analyzes token-level uncertainty across model trajectories. These entropy-sensitive positions serve as natural breakpoints for segmenting the trajectory into semantically meaningful reasoning blocks. Specifically, we define the block structure based on these key points:

\begin{equation}
P_{\text{base}} = B_1 \cup B_2 \cup \dots \cup B_{K+1}
\end{equation}
where each block $ B_n $ contains the subsequence of tokens between two consecutive entropy-sensitive positions:
\begin{equation}
B_n = (t_{k_{n-1}+1}, \dots, t_{k_n}), \quad \text{with } k_0 = 0, \; k_{K+1} = L
\end{equation}

From the previously identified entropy-sensitive positions $ \mathcal{K} $, we derive intermediate reasoning states:
\begin{equation}
S_j = (q, B_1, B_2, \dots, B_j)
\end{equation}
representing the full reasoning state up to and including block $ B_j $, just prior to the entropy-sensitive position $ k_j $.

This entropy-based segmentation provides structure for localized policy refinement and targeted credit propagation. It enables fine-grained control over which segments of the trajectory receive higher attention during optimization. By grounding exploration in model-intrinsic uncertainty signals, FR3E facilitates robust discovery of high-reward reasoning paths, particularly in tasks with sparse rewards and complex action spaces.

\subsection{Entropy-Eliciting Explore: Diversified Path Sampling from Anchors}

For each intermediate state $S_j$, we reset the context and generate $M$ diverse reasoning rollouts:
\begin{equation}
\mathcal{Y}_j := \{Y_{j,m}\}_{m=1}^M, \quad Y_{j,m} \sim \pi_\theta(\cdot \mid S_j)
\end{equation}

Each rollout $Y_{j,m}$ consists of a sequence of generated tokens extending from $S_j$. We define its reward based on whether the generated answer is correct:
\begin{equation}
r_{j,m} = 
\begin{cases}
1 & \text{if } Y_{j,m} \text{ is correct}, \\
0 & \text{otherwise}.
\end{cases}
\end{equation}

Using this reward definition, we estimate the empirical value of $S_j$~\citep{kazemnejad2024vineppounlockingrlpotential} as:
\begin{equation}
V(S_j) = \frac{1}{M} \sum_{m=1}^{M} r_{j,m}
\end{equation}

This value quantifies the average performance gain achieved when transitioning from the prefix $S_j$ to its extensions. It serves as a signal for adaptive policy updates during the Entropy-Eliciting Explore phase.

Through this mechanism, the model effectively samples from local neighborhoods around high-uncertainty prefixes, promoting diversity while preserving semantic coherence with the base trajectory. Specifically, by evaluating the correctness of each rollout, we can identify promising directions for exploration that are likely to lead to correct answers, thereby enhancing both the efficiency and effectiveness of the search process.

\subsection{Adaptive Advantage Modulation for Stable Learning}
\label{subsec:adaptive_advantage}

The phase of Entropy-Eliciting Explore generates a diverse set of rollouts from each intermediate state $S_j$, enabling precise estimation of the corresponding state values $V(S_j)$. To make effective use of this rich, structured reward signal for stable policy learning, we propose an adaptive mechanism that dynamically modulates the advantage function.

We introduce an advantage modulation factor, denoted $\alpha_j$, which scales the learning signal based on the marginal improvement in value from the previous state $S_{j-1}$ to the current state $S_j$:
\begin{equation}
\alpha_j = \frac{1}{\exp\left(V(S_j) - V(S_{j-1})\right)}.
\end{equation}
Using this factor, the modulated advantage $A'(S_j, P_{j,m})$ for a path $P_{j,m}$ originating at state $S_j$ is defined as:
\begin{equation}
A'(S_j, P_{j,m}) = \alpha_j \cdot A(S_j, P_{j,m}).
\end{equation}

The advantage modulation factor functions as a dynamic feedback controller during training. When $V(S_j) > V(S_{j-1})$, indicating a positive progression in reasoning quality, $\alpha_j < 1$. This reduces the influence of the advantage signal for successful paths, preventing premature convergence and preserving exploration diversity. Conversely, when $V(S_j) \leq V(S_{j-1})$, signaling stagnation or degradation in performance, $\alpha_j \geq 1$. This amplifies the learning signal, encouraging more aggressive exploration to overcome reasoning bottlenecks.

Under the idealized assumption that all trajectories within a block have identical length, the average of the modulated advantage across the entire batch is exactly zero:
\begin{equation}
\bar{A}' = \frac{1}{|\mathcal{T}|} \sum_j \alpha_j \sum_{(j, m) \in O_j} (r_{j, m} - V(S_j)) = 0,
\end{equation}
where $|\mathcal{T}|$ denotes the total number of tokens in the batch, and $V(S_j)$ is the empirical estimate of the value of state $S_j$.

Given that each trajectory in block $j$ contributes exactly $L_j$ tokens, the inner sum simplifies to:
\begin{equation}
\sum_{(j, m) \in O_j} (r_{j, m} - V(S_j)) = L_j \sum_{m=1}^{M} (r_{j, m} - V(S_j)),
\end{equation}
which equals zero by the definition of the empirical value. As a result, each term in the outer summation vanishes, leading to $\bar{A}' = 0$.

In practice, minor variations in trajectory lengths and other implementation details may cause deviations from this ideal condition. Nonetheless, the resulting policy gradient estimator remains approximately unbiased and theoretically grounded, while effectively incorporating the dense feedback obtained through structured exploration.

%% file: sections/experiments.tex
\section{Experiments}

%\subsection{Hello World}
\subsection{Setup}

\textbf{Data Source:}
To account for the difficulty distribution of reasoning tasks, we construct our training data from two sources. To ensure stable learning and provide foundational reasoning signals, we use the DeepScaler~\citep{deepscaler2025}dataset as low-difficulty data. For robust and complex reasoning capabilities, we extract samples rated at difficulty levels 3–5 from the SimpleRL~\citep{zeng2025simplerlzooinvestigatingtamingzero} dataset as high-difficulty data. The final training set is formed by combining these two sources to balance stability and challenge throughout training (see Appendix~\ref{app:data_setting} for justification).

\textbf{Training: }
We conduct experiments using the VeRL~\citep{Sheng_2025} framework for reinforcement learning with LLMs. The training setup includes a batch size of 512, a learning rate of $1 \times 10^{-6}$, and a clip range between 0.22 and 0.28. Each response sequence is up to 16k tokens in length. We perform 16 rollouts per prompt and do not apply KL divergence or entropy regularization ($\text{KL Coeff}=0$, $\text{entropy loss}=0$). The mini-batch size is set to 128. Due to the use of rejection sampling, which may result in incomplete batches, we adopt a strategy of accumulating samples until a full batch is formed before proceeding with the training step.

\textbf{Benchmark}: We evaluate performance on standard mathematical reasoning benchmarks, including GSM8K, Math500, Minerva Math, Gaokao2023en, OlympiadBench, College Math and AIME24. For all benchmarks except AIME24, we use greedy sampling for evaluation. Notably, AIME24 employs an \textit{avg@32} strategy, averaging predictions over 32 sampled rollouts per problem to ensure robust performance assessment.
% eval分数的图：base上限提高一些；math相差不大；base32B提升3个点左右
% all wrong的图：均呈下降趋势
% all right的图：均呈上升后逐步稳定的趋势
% response length的图：base后期突然上升，表明开始输出废话；math稳定上升，response在增长；base32B，两种方法都稳定上升
% advantage的图：base先上升，后开始有下降趋势，处在-0.1～0之间；math收敛在0附近；base32B和base7B走势一样
% Entropy loss的图：base先下降，后上升，比GRPO++要上升的早；math也一样；base32B下和GRPO++走势基本一致
\subsection{Main Results}

We analyze the training dynamics of FR3E and GRPO++ across multiple Qwen2.5 variants~\citep{qwen2025qwen25technicalreport}. GRPO++ is vanilla GRPO Integrated with rejection sampling and clip-higher. It is worth noting that alignment of the group size (i.e., number of trajectories per prompt) with standard baseline is inherently infeasible in our setting. Even if we matched the number of rollouts, the inference cost would still differ significantly due to our partial rollout mechanism. We generate only from intermediate states rather than performing full-sequence rollouts from prompts. This design choice enables more efficient exploration and better credit assignment, at the cost of a different computational footprint.

\begin{figure}[htbp]
    \centering
    \begin{subfigure}[b]{0.3\textwidth}
        \includegraphics[width=\textwidth]{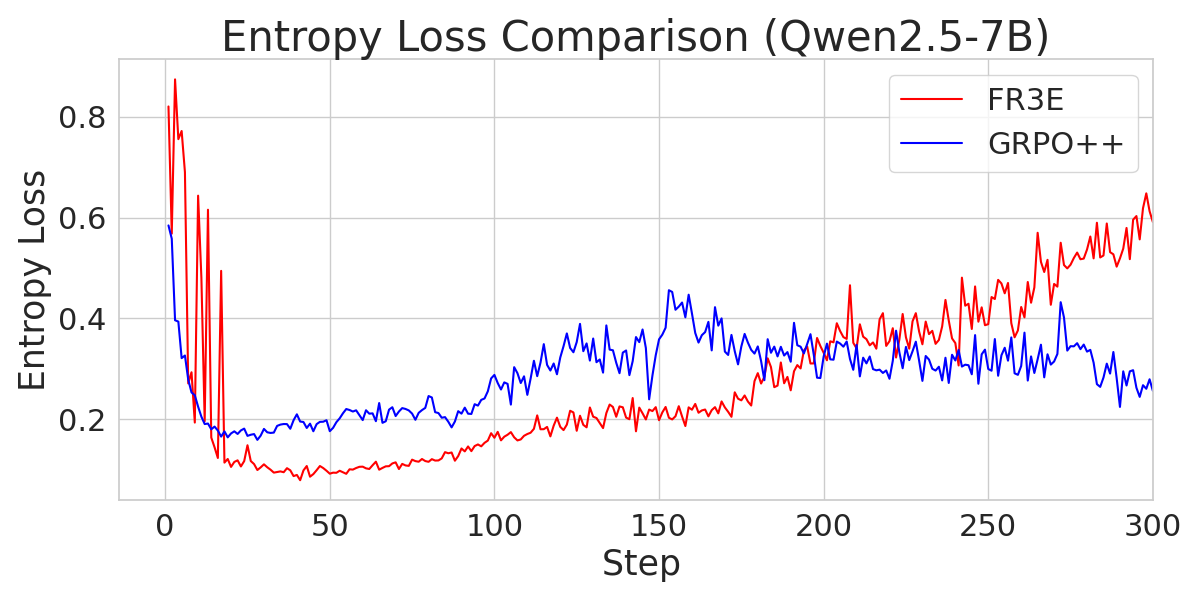}
        \caption{Qwen2.5-7B}
        \label{fig:entropy_base7b}
    \end{subfigure}
    \hfill 
    \begin{subfigure}[b]{0.3\textwidth}
        \includegraphics[width=\textwidth]{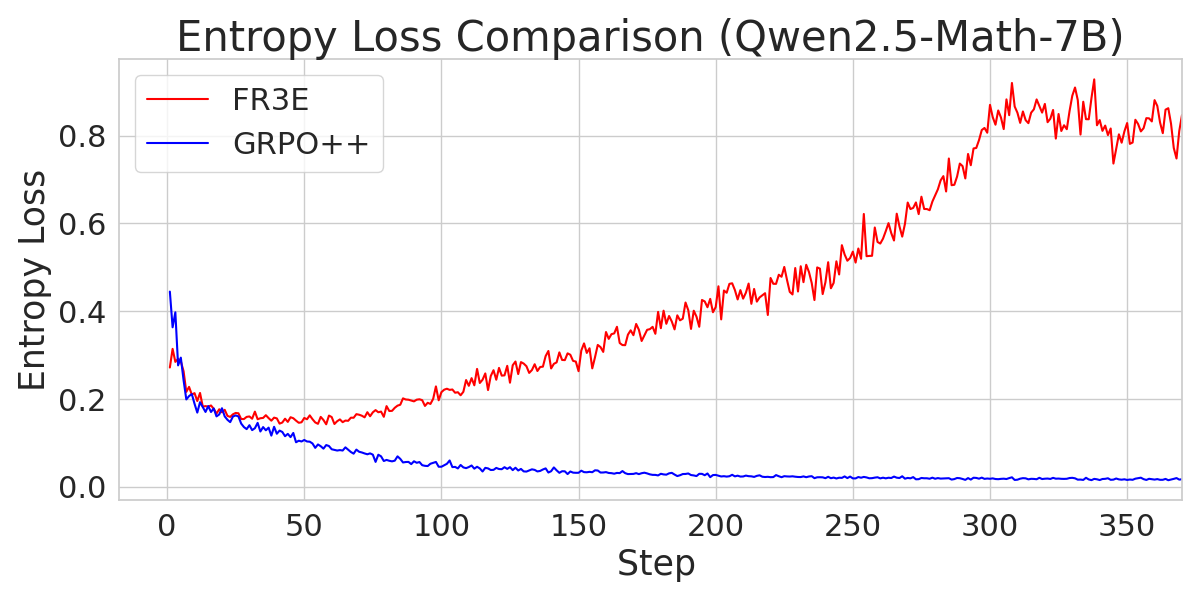}
        \caption{Qwen2.5-Math-7B}
        \label{fig:entropy_math7b}
    \end{subfigure}
    \hfill
    \begin{subfigure}[b]{0.3\textwidth}
        \includegraphics[width=\textwidth]{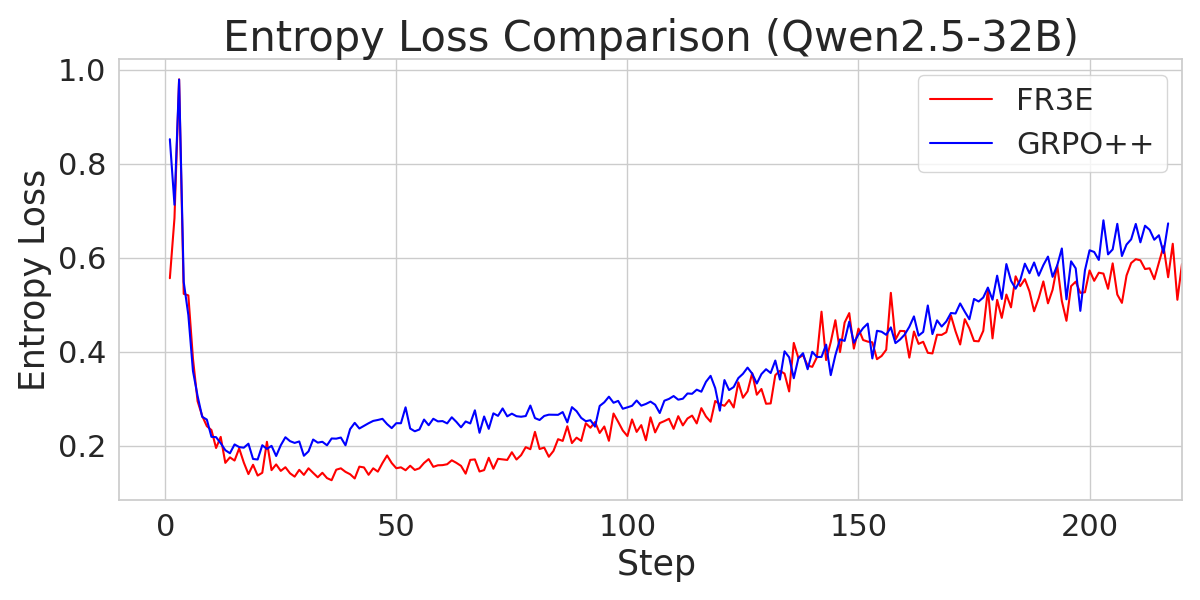}
        \caption{Qwen2.5-32B}
        \label{fig:entropy_base32b}
    \end{subfigure}
    \caption{Entropy Loss Comparison Across Models}
    \label{fig:entropy_loss}
\end{figure}

\begin{figure}[htbp]
    \centering
    \begin{subfigure}[b]{0.3\textwidth}
        \includegraphics[width=\textwidth]{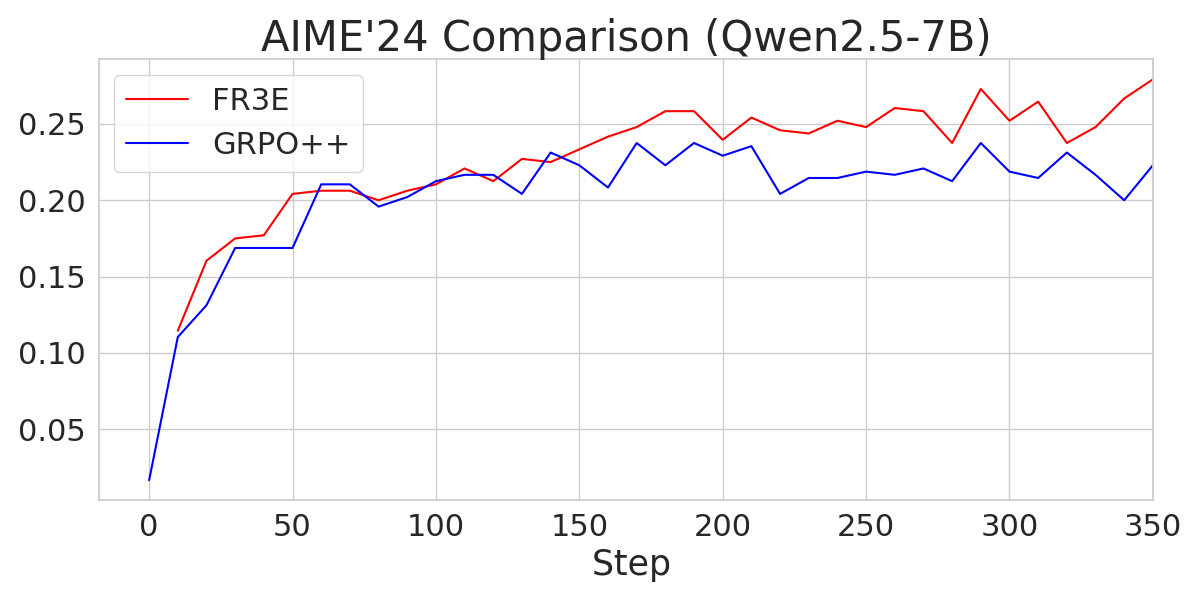}
        \caption{Qwen2.5-7B}
        \label{fig:aime_base7b}
    \end{subfigure}
    \hfill 
    \begin{subfigure}[b]{0.3\textwidth}
        \includegraphics[width=\textwidth]{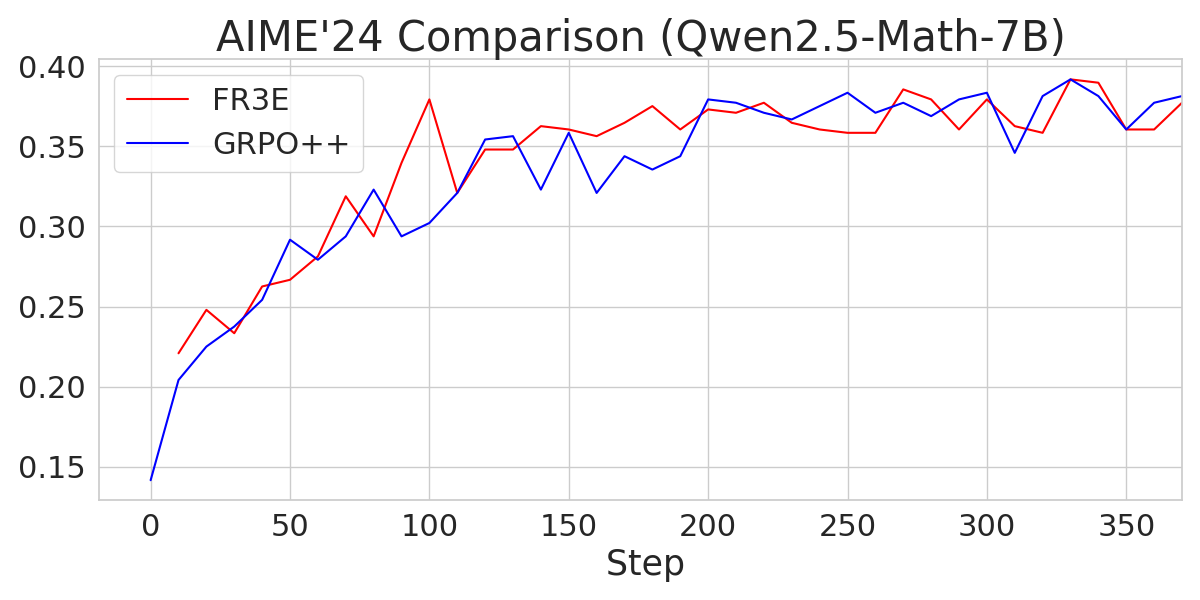}
        \caption{Qwen2.5-Math-7B}
        \label{fig:aime_math7b}
    \end{subfigure}
    \hfill
    \begin{subfigure}[b]{0.3\textwidth}
        \includegraphics[width=\textwidth]{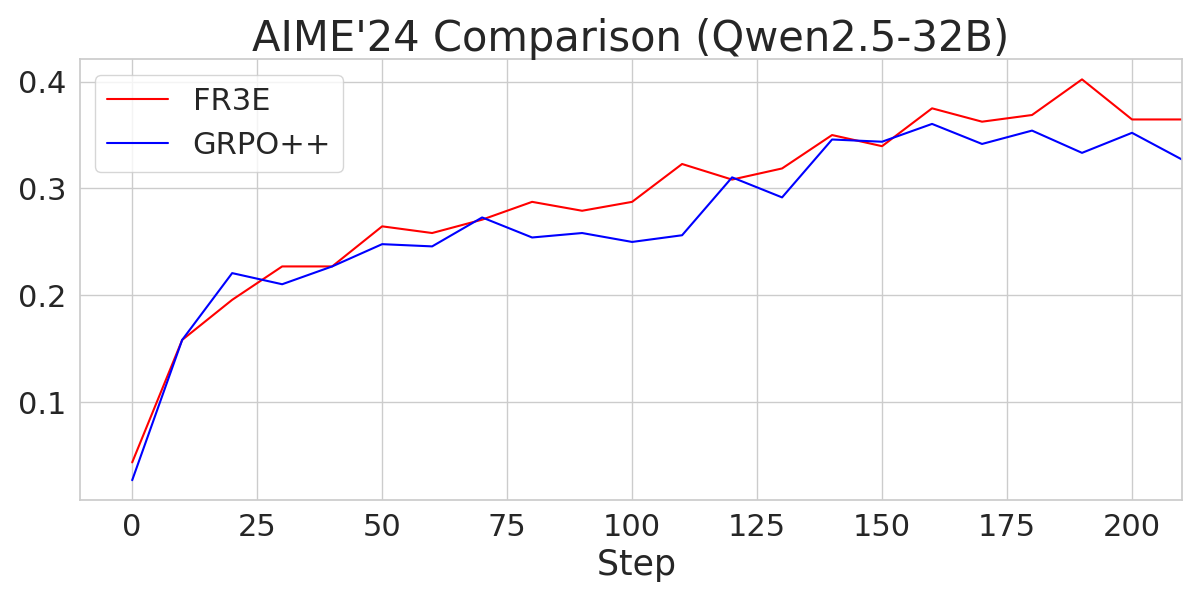}
        \caption{Qwen2.5-32B}
        \label{fig:aime_base32b}
    \end{subfigure}
    \caption{AIME24 Comparison Across Different Models}
    \label{AIME24}
\end{figure}

Figure~\ref{fig:entropy_loss} presents the entropy loss curves for three model variants under both FR3E and GRPO++ training, revealing distinct patterns in exploration behavior and learning dynamics. In general, all models exhibit an initial sharp decline in entropy, indicating a strong tendency toward exploitation during early training stages. However, differences emerge as training progresses. For Qwen2.5-7B and Qwen2.5-32B, FR3E maintains slightly higher entropy levels compared to GRPO++, suggesting a more balanced trade-off between exploration and exploitation. The most striking observation is seen in Qwen2.5-Math-7B, where FR3E exhibits a gradual increase in entropy during later stages of training — a contrast to the rapid convergence to low entropy seen in GRPO++. This indicates that FR3E preserves exploration capacity longer, which may be especially advantageous in complex reasoning tasks that benefit from diverse solution paths. These results collectively highlight how exploration strategy influences not only learning stability but also exploration behavior, particularly in domain-specialized models.

Figure~\ref{AIME24} shows that FR3E achieves higher performance than GRPO++ under the same training steps, especially for Qwen2.5-7B and Qwen2.5-32B model. This indicates that FR3E can effectively push the performance ceiling of larger models. Notably, the GRPO++ baseline on Qwen2.5-32B performs at a level consistent with the M1-reproduced DAPO results, suggesting normal baseline behavior \citep{minimax2025minimaxm1scalingtesttimecompute}. In contrast, on Qwen2.5-Math-7B, the gap between FR3E and GRPO++ is smaller, suggesting limited gains in this specialized setting. Nevertheless, FR3E still maintains slightly more stable learning dynamics.Overall, FR3E demonstrates strong scalability and stability across general-purpose models, with the most significant benefits observed in large-scale architectures.

% \begin{figure}[htbp]
%     \centering
%     \begin{subfigure}[b]{0.3\textwidth}
%         \includegraphics[width=\textwidth]{figure/reward_fr3e_vs_grpo_310_steps.png}
%         \caption{Qwen2.5-7B}
%         \label{fig:reward_base7b}
%     \end{subfigure}
%     \hfill 
%     \begin{subfigure}[b]{0.3\textwidth}
%         \includegraphics[width=\textwidth]{figure/reward_fr3e_vs_grpo_310_steps_math7b.png}
%         \caption{Qwen2.5-Math-7B}
%         \label{fig:reward_math7b}
%     \end{subfigure}
%     \hfill
%     \begin{subfigure}[b]{0.3\textwidth}
%         \includegraphics[width=\textwidth]{figure/reward_fr3e_vs_grpo_310_steps_base32b.png}
%         \caption{Qwen2.5-32B}
%         \label{fig:reward_base32b}
%     \end{subfigure}
%     \caption{Reward Stability Comparsion Across Different Models}
%     \label{Reward}
% \end{figure}

\begin{figure}[htbp]
    \centering
    \begin{subfigure}[b]{0.3\textwidth}
        \includegraphics[width=\textwidth]{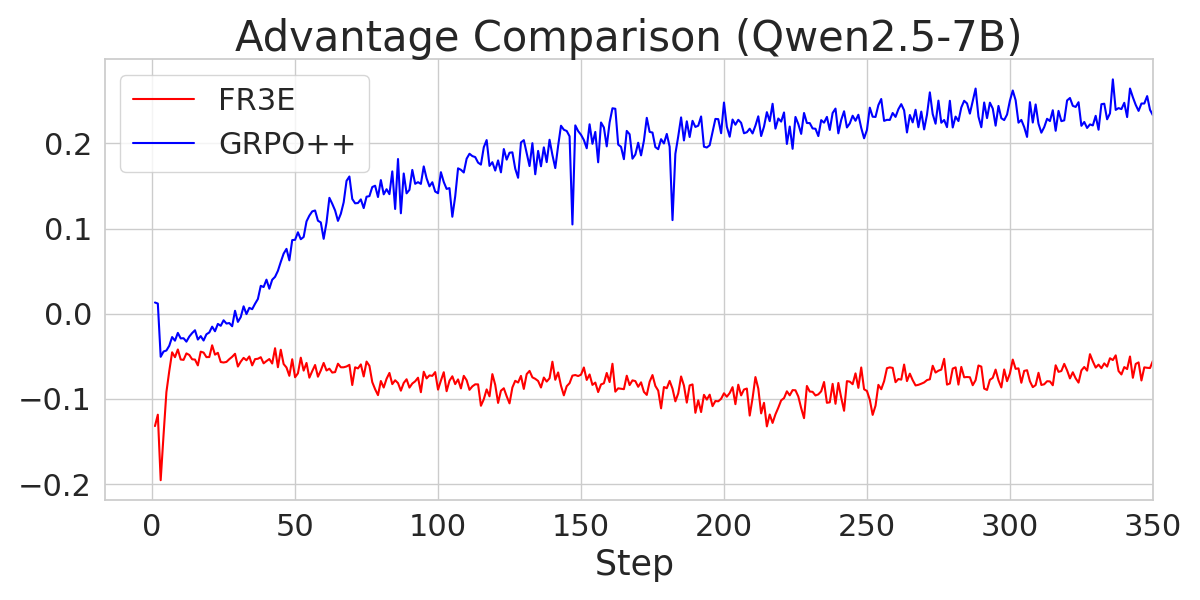}
        \caption{Qwen2.5-7B}
        \label{fig:sub1}
    \end{subfigure}
    \hfill 
    \begin{subfigure}[b]{0.3\textwidth}
        \includegraphics[width=\textwidth]{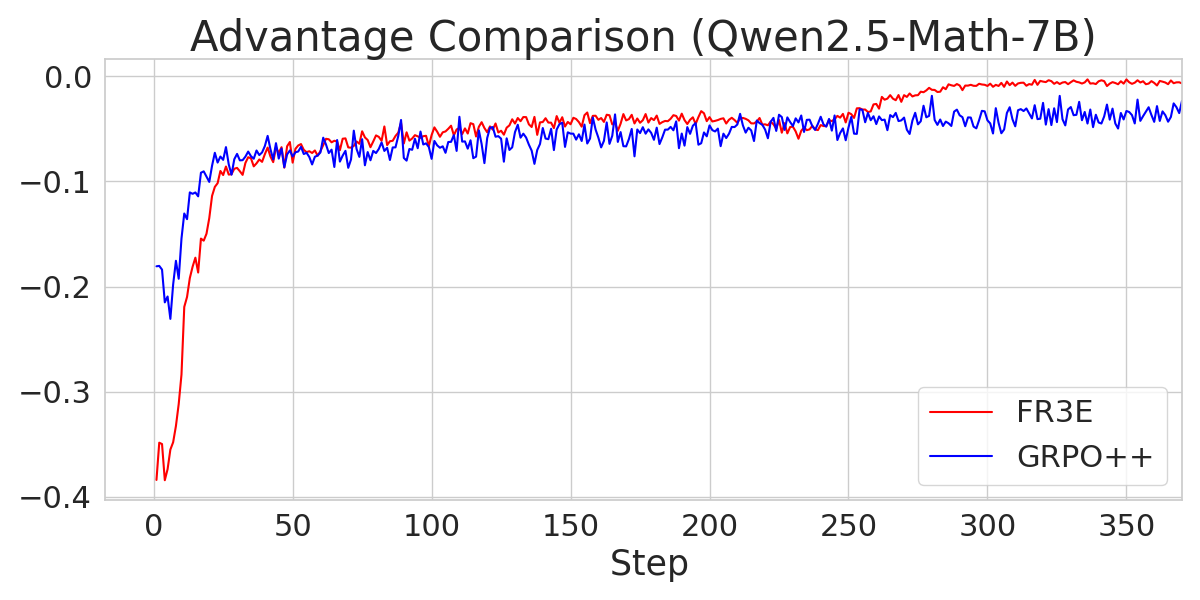}
        \caption{Qwen2.5-Math-7B}
        \label{fig:sub2}
    \end{subfigure}
    \hfill
    \begin{subfigure}[b]{0.3\textwidth}
        \includegraphics[width=\textwidth]{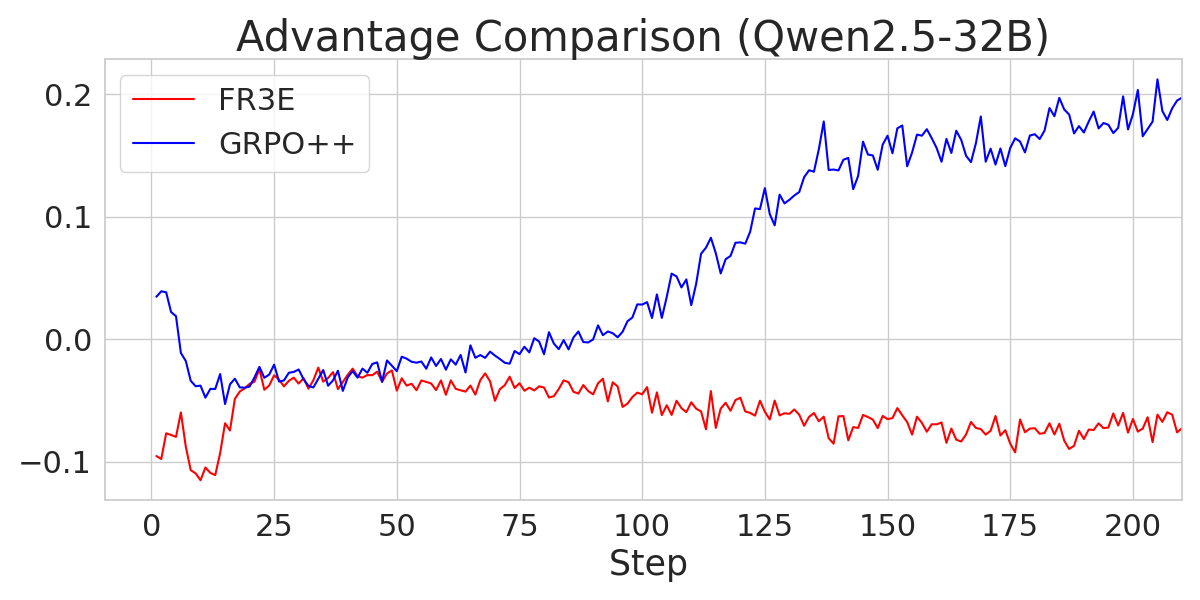}
        \caption{Qwen2.5-32B}
        \label{fig:sub3}
    \end{subfigure}
    \caption{Advantage Comparison Across Different Models}
    \label{Advantage}
\end{figure}

Our method generates more stable advantage estimates compared to GRPO++. As illustrated in Figure~\ref{Advantage}, the advantage values exhibit smaller fluctuations and remain tightly centered around zero throughout training. The theoretical expectation $\mathbb{E}_{a \sim \pi(\cdot|s)}[A(s,a)] = 0$ is closely preserved in our approach, suggesting minimal distributional shift between the current policy and the data collection process.

\begin{table}[htbp]
\centering
\caption{Multi-Benchmark comparison (Accuracy \%)}
\label{tab:performance}
\rowcolors{2}{lightblue}{white}
\begin{tabular}{l*{6}{c}}
\toprule
\rowcolor{headerblue}
\multicolumn{1}{c}{} & 
\multicolumn{2}{c}{\textbf{Qwen2.5-Math-7B}} & 
\multicolumn{2}{c}{\textbf{Qwen2.5-7B}} & 
\multicolumn{2}{c}{\textbf{Qwen2.5-32B}} \\
\rowcolor{headerblue}
 \multicolumn{1}{l}{\textbf{Benchmark}} & \textbf{FR3E} & \textbf{GRPO++} & \textbf{FR3E} & \textbf{GRPO++} & \textbf{FR3E} & \textbf{GRPO++} \\
\midrule
AIME24        & 39.1(+0.0\%) & 39.1 & 25.2(+2.5\%) & 22.7 & 40.2(+6.1\%) & 34.1 \\
GSM8k         & 91.3(+0.1\%) & 91.2 & 92.8(+1.6\%) & 91.2 & 96.1(+0.3\%) & 95.8 \\
Math500       & 82.2(+0.6\%) & 81.6 & 79.0(+1.2\%) & 77.8 & 87.4(+2.2\%) & 85.2 \\
Minerva Math  & 40.8(+2.6\%) & 38.2 & 39.0(+3.7\%) & 35.3 & 45.6(+2.6\%) & 43.0 \\
Gaokao2023en  & 67.8(+2.6\%) & 65.2 & 67.3(+3.4\%) & 63.9 & 75.3(+3.9\%) & 71.4 \\
Olympiadbench & 46.5(+3.5\%) & 43.0 & 42.1(+3.3\%) & 38.8 & 51.7(+3.0\%) & 48.7 \\
College Math  & 47.4(-0.1\%) & 47.5 & 45.1(+0.0\%) & 45.1 & 48.3(+1.3\%) & 47.0 \\
AMC23         & 67.5(+5.0\%) & 62.5 & 67.5(+7.5\%) & 60.0 & 80.0(+5.0\%) & 75.0 \\
Avg           & 60.3(+1.8\%) & 58.5 & 57.3(+3.0\%) & 54.3 & 65.6(+3.1\%) & 62.5 \\
\bottomrule
\end{tabular}

\vspace{0.2cm}
\end{table}

From the multi-benchmark comparison shown in Table~\ref{tab:performance}, we observe that FR3E demonstrates strong generalization capabilities across various reasoning tasks, particularly when applied to general-purpose models. Notably, FR3E consistently achieves competitive results, often matching or slightly outperforming GRPO++. 

Interestingly, FR3E exhibits limited gains on the domain-specialized Qwen2.5-Math-7B model, where it performs comparably to GRPO++ across most benchmarks. This suggests that applying standard RL strategies to a math-specialized model may not yield significant improvements, possibly due to interference with its fine-tuned knowledge.

In contrast, when RL is applied to the base Qwen2.5-7B model, FR3E delivers substantial improvements across nearly all benchmarks. Similar trends are observed on the larger Qwen2.5-32B backbone, where FR3E consistently outperforms GRPO++. These results highlight the importance of aligning model specialization with appropriate training methodologies — structured reinforcement learning proves highly effective for generalist models, but may require more careful adaptation for domain-specific ones~\citep{shao2025spurious}.

\newpage

\subsection{Discussion}

\subsubsection{Training Dynamics of Qwen-Math-7B}
\label{subsec:qwen_math_rl_issue}
\begin{figure}[htbp]
    \centering
    \begin{subfigure}[b]{0.3\textwidth}
        \includegraphics[width=\textwidth]{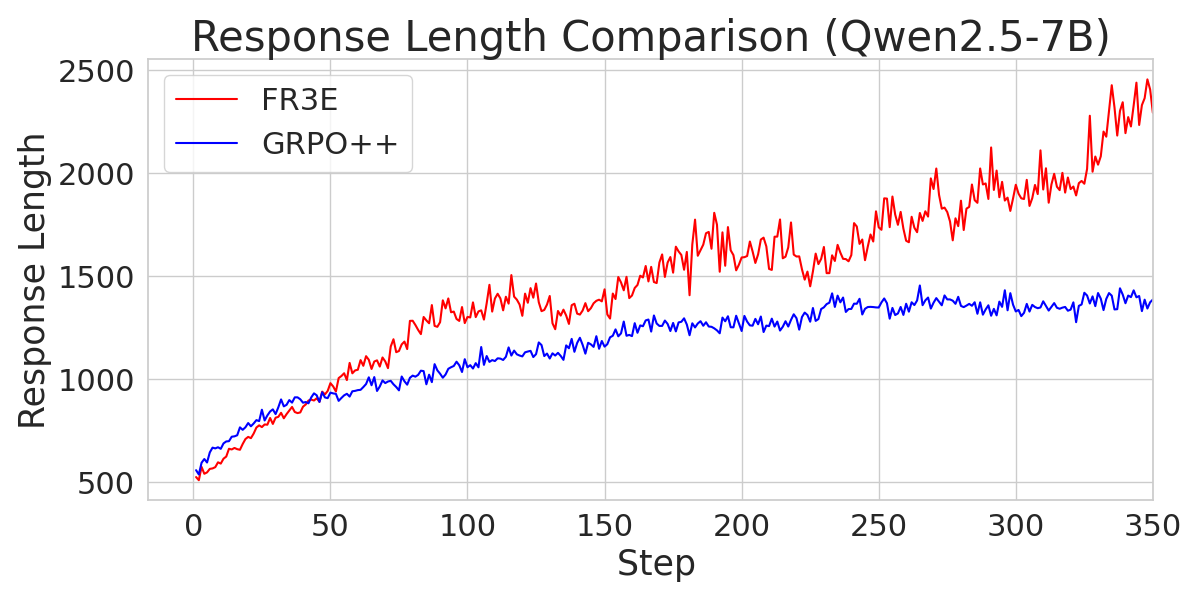}
        \caption{Qwen2.5-7B}
        \label{fig:response_base7b}
    \end{subfigure}
    \hfill 
    \begin{subfigure}[b]{0.3\textwidth}
        \includegraphics[width=\textwidth]{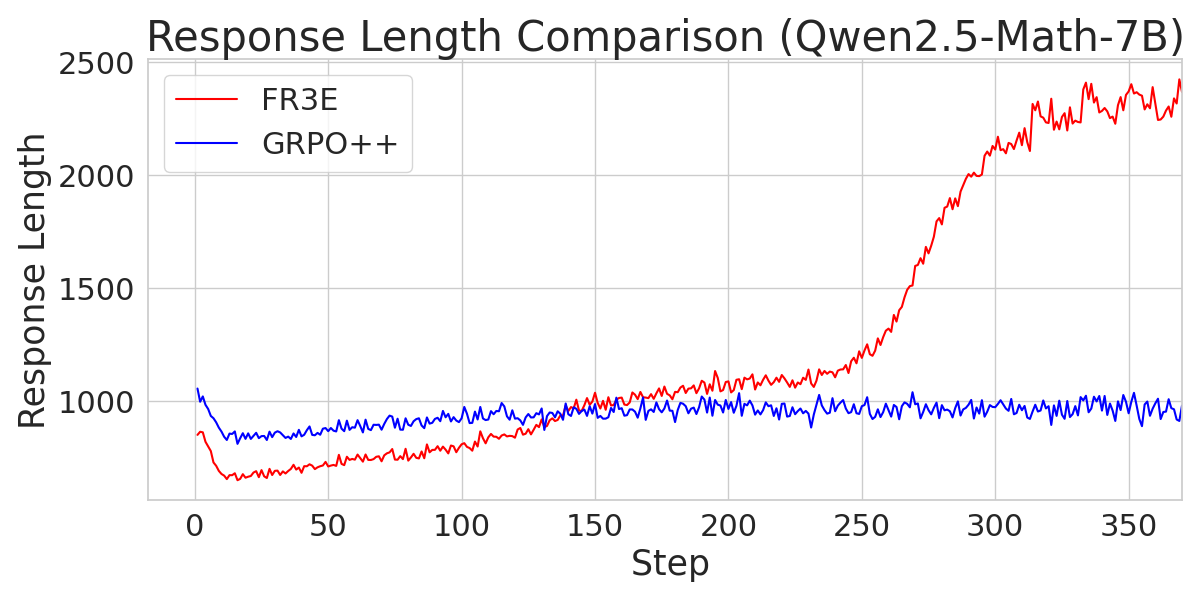}
        \caption{Qwen2.5-Math-7B}
        \label{fig:response_math7b}
    \end{subfigure}
    \hfill
    \begin{subfigure}[b]{0.3\textwidth}
        \includegraphics[width=\textwidth]{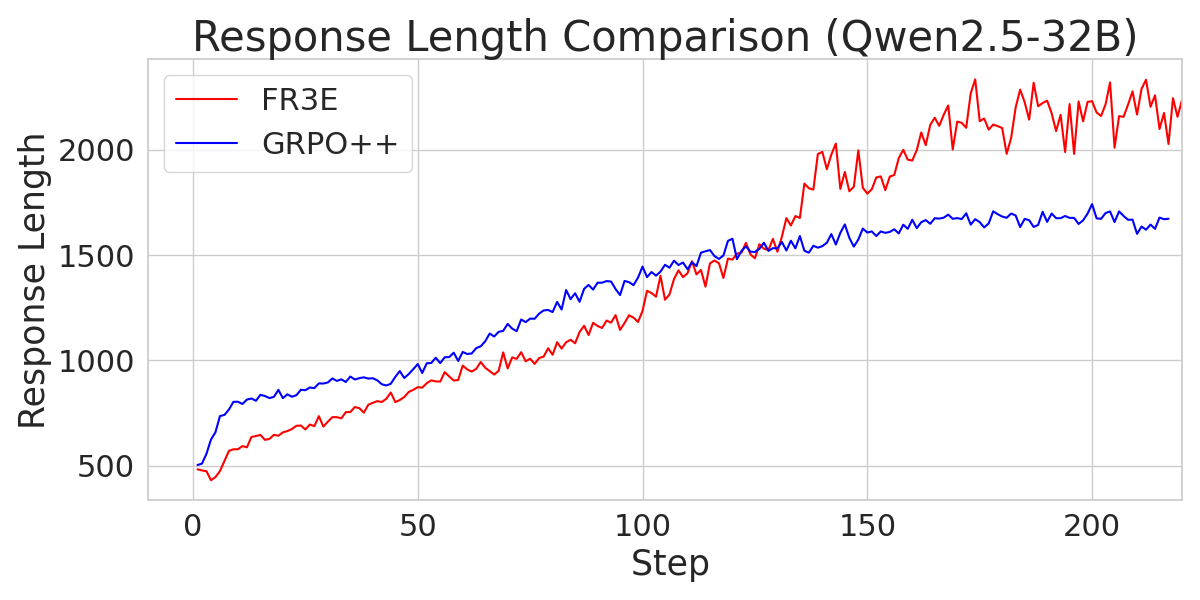}
        \caption{Qwen2.5-32B}
        \label{fig:response_base32b}
    \end{subfigure}
    \caption{Response Length Comparison Across Different Models}
    \label{Response Length}
\end{figure}
As noted in the Main Results, the RL performance of Qwen2.5-Math-7B exhibits distinct patterns compared to other models, motivating a closer examination of its training dynamics.

We observe several notable characteristics during training:

\begin{itemize}
    \item \textbf{Early saturation in response length:} The model quickly reaches the maximum sequence length limit, which may constrain the development of longer reasoning paths.
    \item \textbf{Overstable entropy trends:} Compared to more stable entropy decay observed in general-purpose models, Qwen2.5-Math-7B shows greater variability and earlier drops in entropy loss, suggesting potentially reduced exploration capacity during training.
\end{itemize}

% \begin{figure}[htbp]
%     \centering
%     \includegraphics[width=0.9\textwidth]{figure/performance_multi_bench.png} 
%     \caption{Multi-benchmark performance comparison across different models.}
%     \label{fig:multi-bench}
% \end{figure}

When comparing training algorithms, we observe differences in how Qwen2.5-Math-7B adapts to GRPO++ and FR3E. As shown in Figure~\ref{fig:response_math7b}, the average response length remains relatively flat under GRPO++, while it increases steadily under FR3E. This suggests that FR3E’s trajectory-level reward shaping may better support the development of extended reasoning sequences for this model.

A similar pattern is reflected in the AIME24 (Figure~\ref{fig:aime_math7b}, where GRPO++ yields fluctuating and less stable performance, whereas FR3E demonstrates a more consistent upward trend. While Qwen2.5-Math-7B shows unique training dynamics under standard RL methods, FR3E appears to offer certain advantages in promoting structured reasoning and stabilizing learning progress. 

\subsubsection{Sources of Gains in FR3E}

To better understand the source of FR3E's performance improvement, we analyze it from multiple perspectives.

\textbf{Higher Entropy Enable Healthier Exploration} The entropy loss trends across different model sizes provide significant insights into the exploration-exploitation dynamics of FR3E compared to GRPO++. Initially, both algorithms exhibit a sharp decrease in entropy loss (Figure~\ref{fig:entropy_base7b}), indicating rapid reduction in policy uncertainty during the early stages of training on the Qwen2.5-7B model. However, as training progresses, FR3E demonstrates a higher entropy loss than GRPO++, suggesting that it retains greater exploration capability and avoids premature convergence. This trend is also observed in a different scale setting (Figure~\ref{fig:entropy_math7b}), where FR3E shows a steady increase in entropy loss, indicative of its stable exploration strategy. Conversely, GRPO++ converges more quickly to a lower entropy value, which might limit its ability to explore diverse solutions effectively. On the larger Qwen2.5-32B model (Figure~\ref{fig:entropy_base32b}), while both algorithms start with similar initial behavior, FR3E maintains a gradual increase in entropy loss, highlighting its effectiveness in balancing exploration and exploitation. In contrast, the GRPO++'s rapid convergence to a lower entropy suggests reduced exploration efficiency.

Notably, in the Qwen2.5-Math-7B variant, the entropy loss stabilizes at a relatively low level much earlier in training, yet the final AIME24 score remains comparable to that of models with higher entropy trajectories. This apparent discrepancy suggests that while entropy is an important indicator of exploration, it may not be the sole determinant of final performance, especially for domain-specific models that may encode stronger prior knowledge but lack the flexibility to adapt more dynamical pattern during RL fine-tuning. The early saturation of entropy loss in Qwen2.5-Math-7B, coupled with its limited maximum sequence length, likely constrains its ability to generate longer, more complex reasoning paths, thereby limiting its full potential under standard RL frameworks.

These observations collectively indicate that FR3E to maintaining higher entropy throughout training fosters better exploration, contributing to its superior performance in accuracy and stability. Furthermore, the impact of model size becomes evident, with larger models showing more pronounced differences in entropy loss between the two methods, underscoring the importance of sustained exploration in complex models.

\begin{figure}[htbp]
    \centering
    \begin{subfigure}[b]{0.3\textwidth}
        \includegraphics[width=\textwidth]{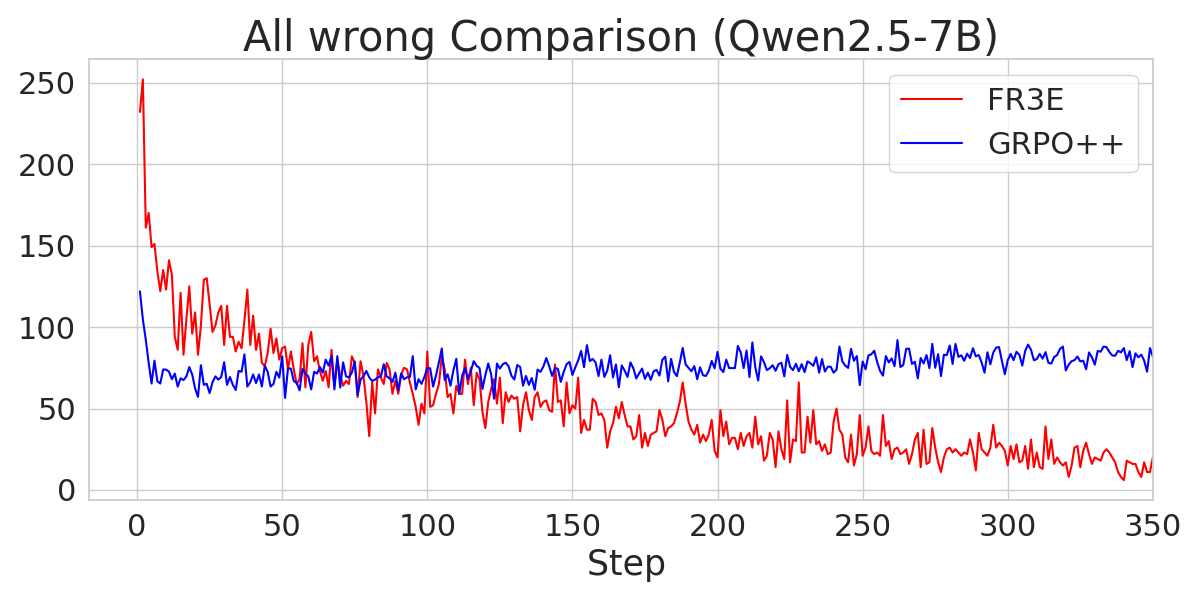}
        \caption{Qwen2.5-7B}
        \label{fig:all-wrong_base7b}
    \end{subfigure}
    \hfill 
    \begin{subfigure}[b]{0.3\textwidth}
        \includegraphics[width=\textwidth]{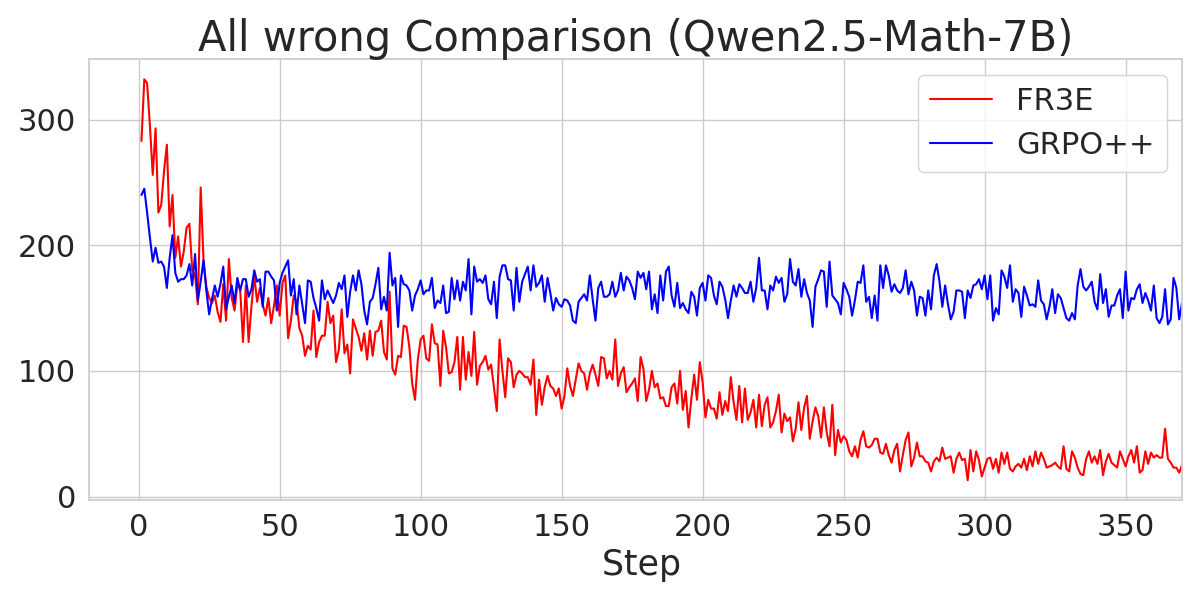}
        \caption{Qwen2.5-Math-7B}
        \label{fig:all-wrong_math7b}
    \end{subfigure}
    \hfill
    \begin{subfigure}[b]{0.3\textwidth}
        \includegraphics[width=\textwidth]{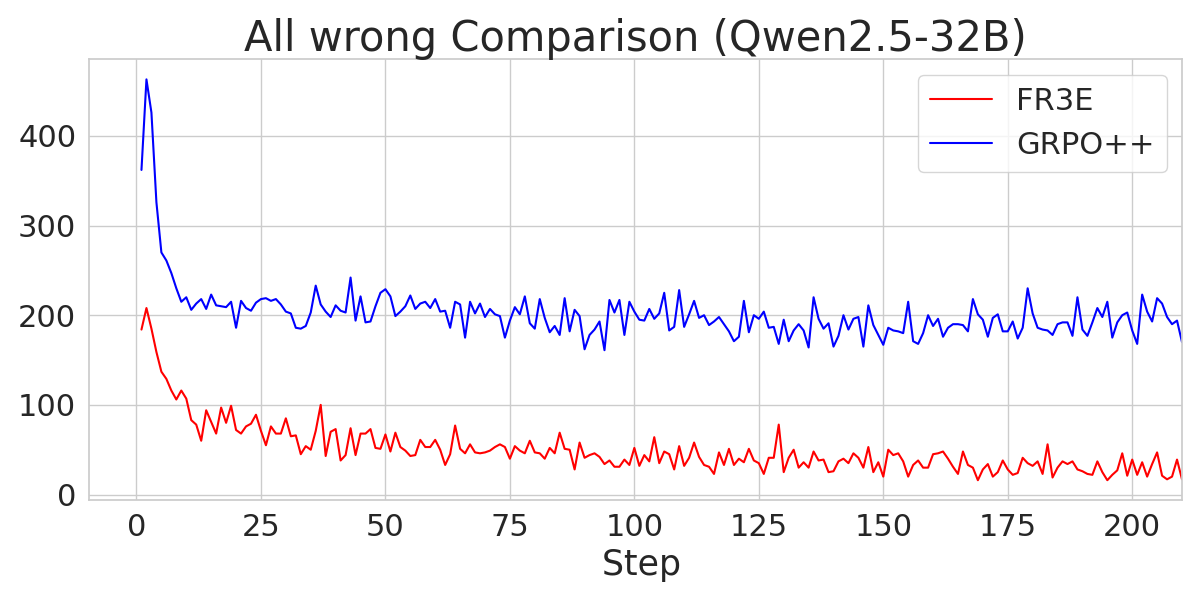}
        \caption{Qwen2.5-32B}
        \label{fig:all-wrong_base32b}
    \end{subfigure}
    \caption{Number of All-Wrong Trajectories during the Entropy-Eliciting Explore Phase}
    \label{Second All Wrong Number}
\end{figure}

\begin{figure}[htbp]
    \centering
    \begin{subfigure}[b]{0.3\textwidth}
        \includegraphics[width=\textwidth]{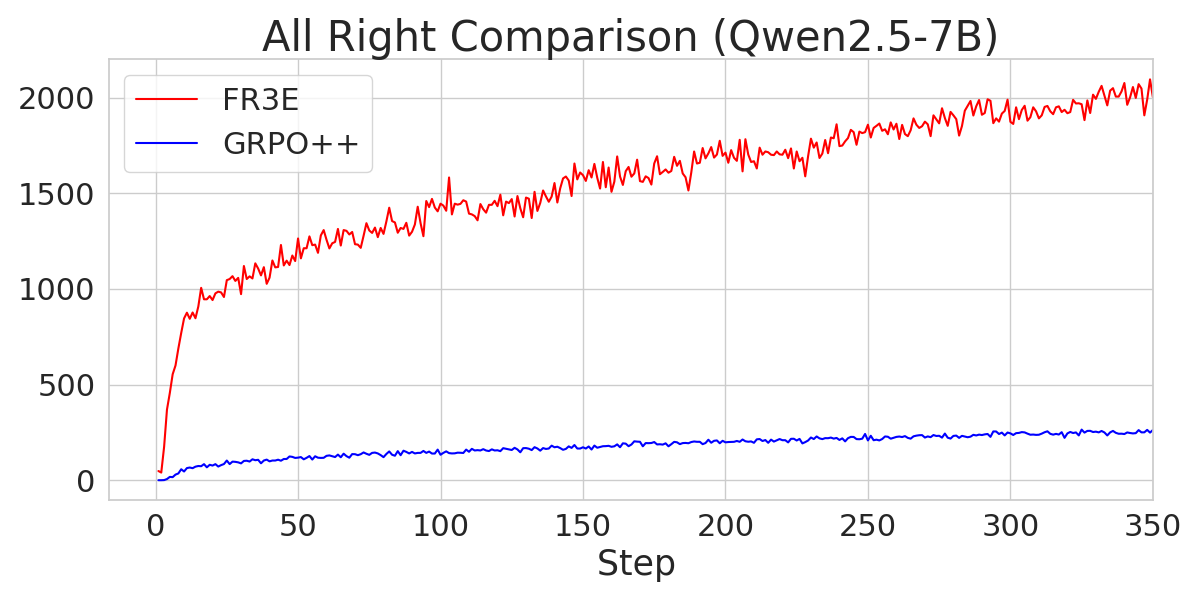}
        \caption{Qwen2.5-7B}
        \label{fig:all-right_base7b}
    \end{subfigure}
    \hfill 
    \begin{subfigure}[b]{0.3\textwidth}
        \includegraphics[width=\textwidth]{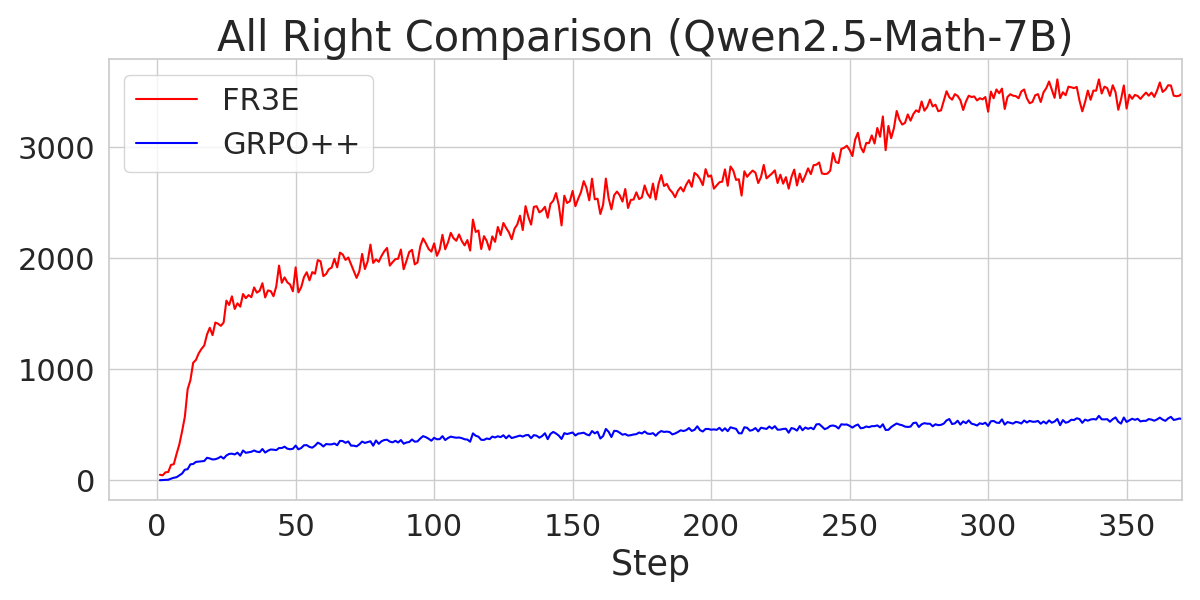}
        \caption{Qwen2.5-Math-7B}
        \label{fig:all-right_math7b}
    \end{subfigure}
    \hfill
    \begin{subfigure}[b]{0.3\textwidth}
        \includegraphics[width=\textwidth]{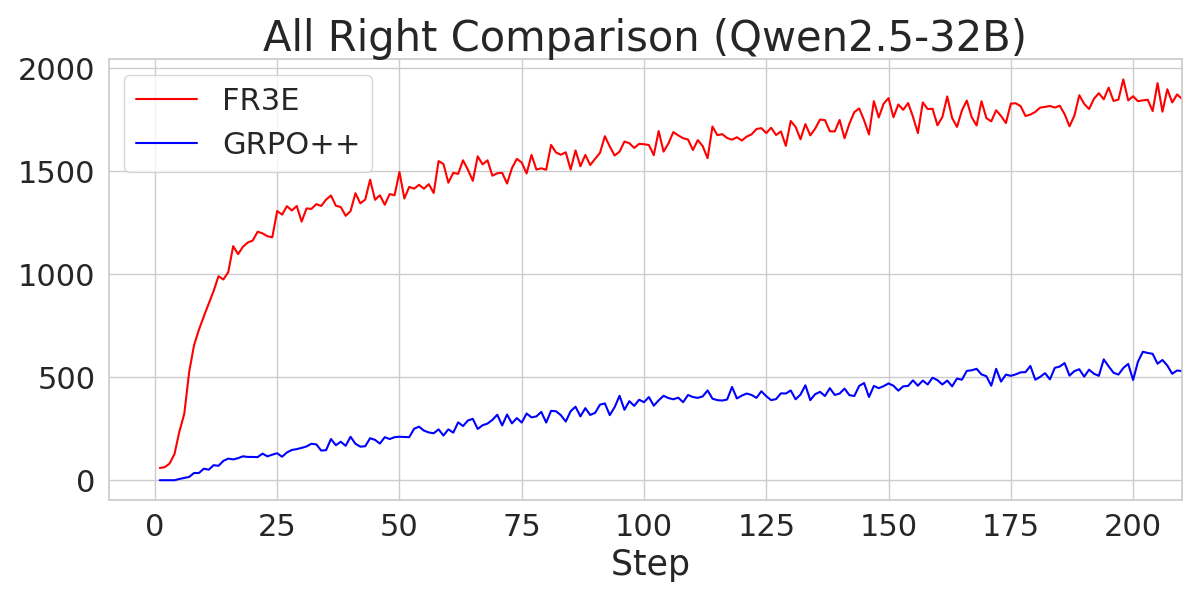}
        \caption{Qwen2.5-32B}
        \label{fig:all-right_base32b}
    \end{subfigure}
    \caption{Number of All-Right Trajectories during the Entropy-Eliciting Explore Phase}
    \label{Second All Right Number}
\end{figure}
\textbf{Entropy-Eliciting Explore: Improving Trajectory Consistency}  We analyze the trajectory-level consistency of exploration by measuring, for each prompt, whether all rollouts are either completely incorrect (``All-Wrong'') or completely correct (``All-Right''). As shown in Figure~\ref{Second All Wrong Number} and Figure~\ref{Second All Right Number}, this analysis reveals two key trends. First, FR3E significantly increases the number of All-Right trajectories over time while suppressing All-Wrong ones. This indicates that the algorithm successfully guides the policy toward generating more consistent and high-quality reasoning paths, rather than relying on noisy or random successes. The increasing gap between All-Right and All-Wrong curves suggests that FR3E accumulates reliable positive signals through its trajectory reweighting mechanism, leading to more stable and meaningful updates. Second, we observe notable differences across model variants. On Qwen2.5-Math-7B (Figure~\ref{fig:all-wrong_math7b} and Figure~\ref{fig:all-right_math7b}), although the model converges early in entropy (as discussed in Subsection~\ref{subsec:qwen_math_rl_issue}), it still achieves a moderate increase in All-Right trajectories, suggesting that domain-specific priors may help bootstrap performance even under limited exploration. In contrast, Qwen2.5-7B (Figure~\ref{fig:all-wrong_base7b} and Figure~\ref{fig:all-right_base7b}) shows faster and more consistent improvements, with a sharper rise in All-Right and sharper drop in All-Wrong trajectories. Finally, the largest model, Qwen2.5-32B (Figure~\ref{fig:all-wrong_base32b} and Figure~\ref{fig:all-right_base32b}), demonstrates the most robust behavior, achieving the highest All-Right count and lowest All-Wrong count, highlighting the benefit of larger capacity in leveraging FR3E's exploration strategy.

These results demonstrate that FR3E enhances the quality of exploration not only by reducing reliance on misleading or sparse rewards but also by promoting consistent, high-quality reasoning paths. This aligns with our hypothesis that trajectory-level reward shaping can lead to more reliable and scalable reinforcement learning in complex reasoning tasks.

\textbf{Test Distribution Analysis} 
\begin{figure}[htbp]
    \centering
    \begin{subfigure}[t]{0.48\textwidth}  
        \includegraphics[width=\linewidth]{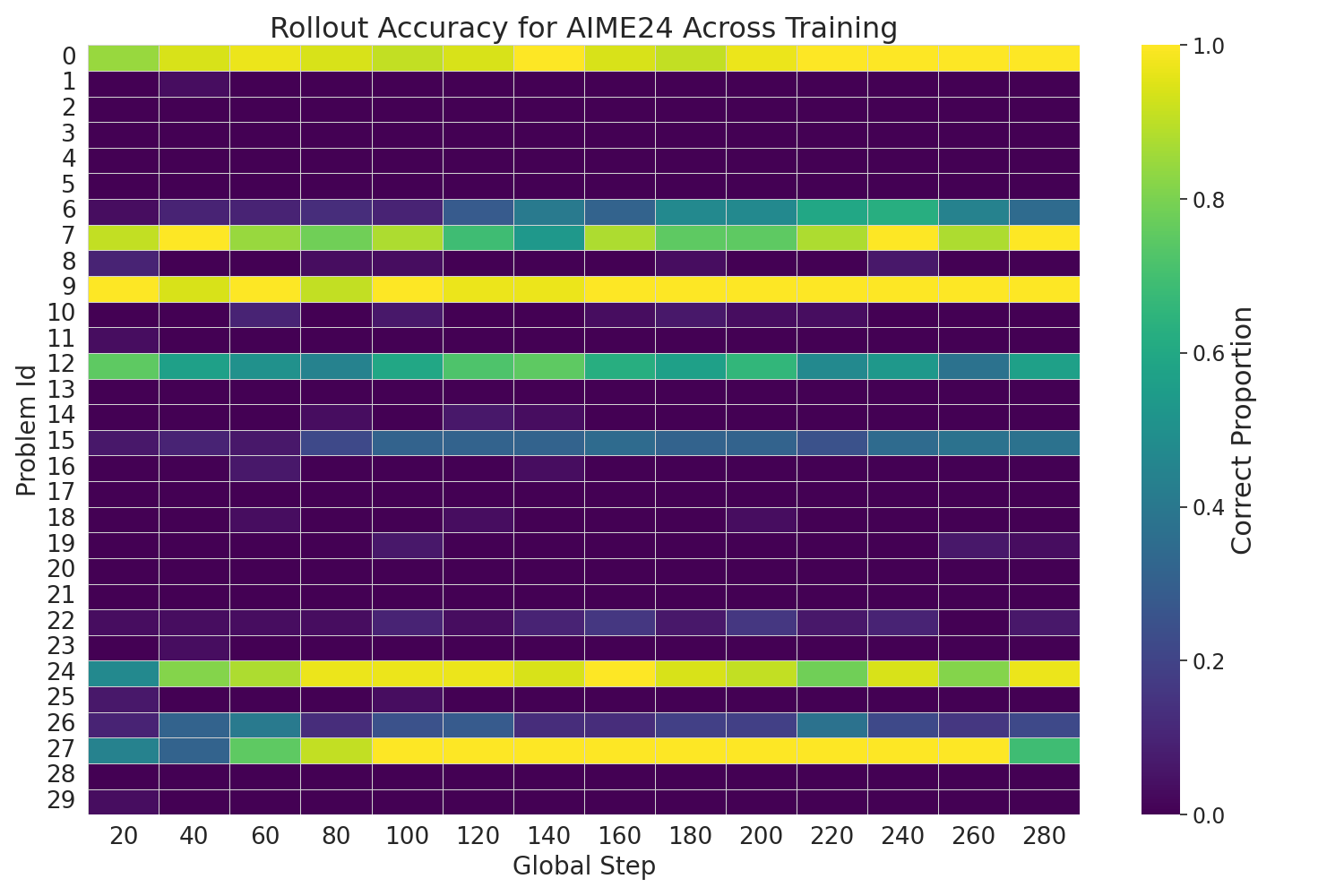}
        \caption{FR3E(Qwen2.5-7B)}
        \label{fig:fr3e-heatmap}
    \end{subfigure}
    \hfill
    \begin{subfigure}[t]{0.48\textwidth}
        \includegraphics[width=\linewidth]{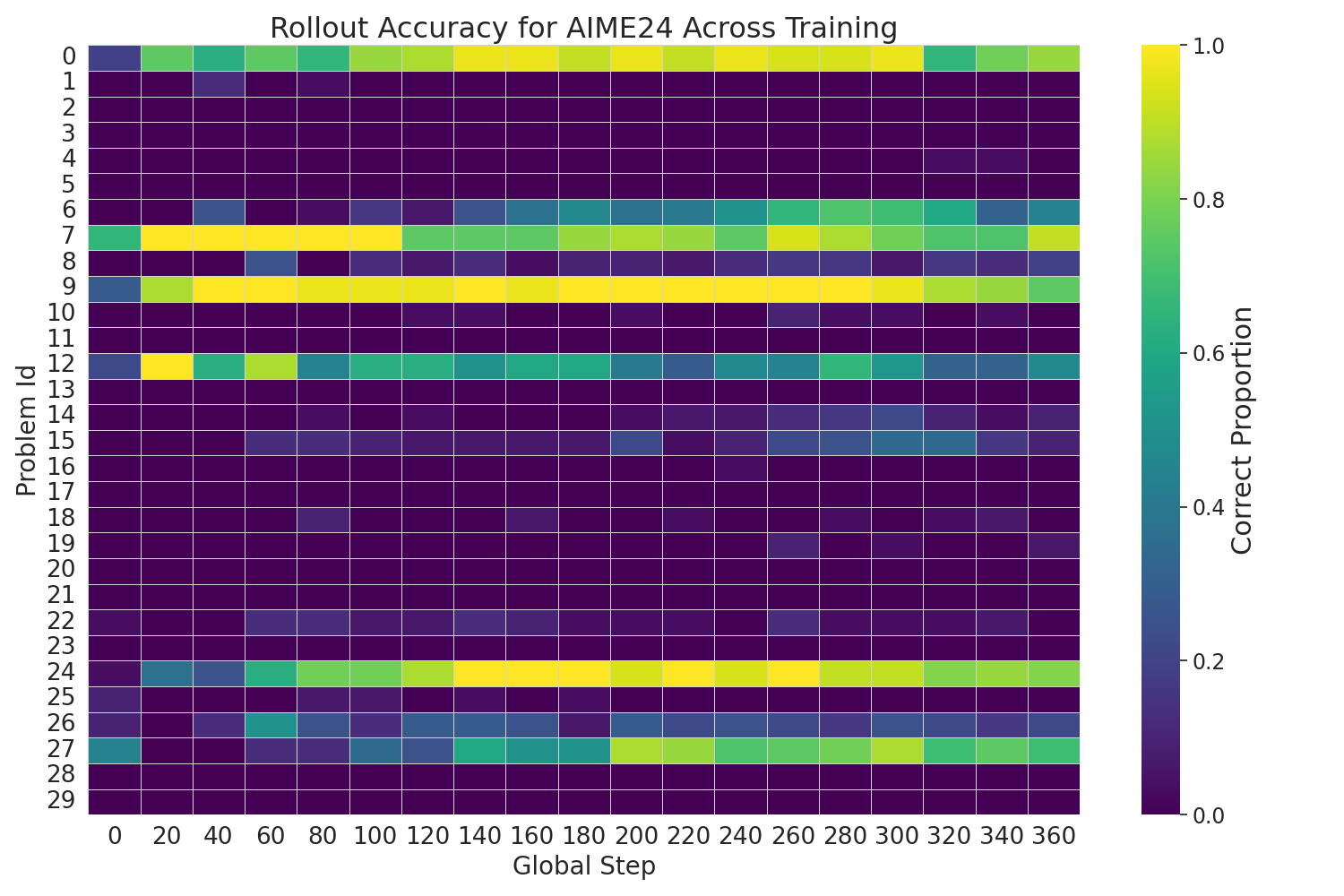}
        \caption{GRPO++(Qwen2.5-7B)}
        \label{fig:GRPO++-heatmap}
    \end{subfigure}
    \caption{Rollout accuracy heatmap comparison (Qwen2.5-7B)}
    \label{fig:rollout-analysis}
\end{figure}

To better understand the learning dynamics of FR3E during training and how it improves reasoning stability over time, we visualize and compare the rollout accuracy across all AIME problems throughout the training process for both FR3E and GRPO++.

As training progresses, the heatmap for FR3E Figure~\ref{fig:fr3e-heatmap} shows a consistent transition from low-accuracy regions to high-accuracy regions. This pattern indicates that FR3E gradually improves its reasoning performance in a stable and progressive manner, rather than relying on sudden or sporadic improvements~\citep{li2025temporalsamplingforgottenreasoning}. Notably, the emergence of sharp peaks in the heatmap, where certain problems quickly converge to consistently high accuracy, suggests that once the model learns to solve a problem correctly, it retains this capability without significant fluctuations. This implies:

\begin{itemize}
    \item \textbf{Stable learning dynamics}: The model does not oscillate between correct and incorrect answers after convergence.
    \item \textbf{High confidence in solutions}: Once a problem is mastered, the model reliably generates correct reasoning chains.
    \item \textbf{Effective suppression of error propagation}: FR3E avoids compounding errors during training, which is critical in long-chain reasoning tasks.
\end{itemize}

In contrast, the heatmap for GRPO++ Figure~\ref{fig:GRPO++-heatmap} exhibits more variability and less consistent improvement. While some problems show rapid convergence, others remain at lower accuracy levels even after extensive training. This inconsistency highlights the potential instability in GRPO++'s learning process.

The training process of FR3E demonstrates a robust and incremental learning pattern, where problem-solving capabilities are acquired gradually and steadily. Crucially, once a problem is learned, the model maintains consistently high performance across multiple rollouts, indicating that our method enables the model to internalize solutions more firmly — effectively "locking in" knowledge with greater reliability than baseline approaches.

% \subsection{Summary}

% FR3E demonstrates consistent advantages on general-purpose base models, particularly in terms of training stability, response length control, and exploration efficiency. Meanwhile, the RL results of Qwen-Math-7B highlight the potential adaptability risks of domain-specific models within structured reinforcement learning settings. Future work should seek finer-grained alignment between algorithm design and model architecture to achieve more efficient reasoning enhancement.

%% file: sections/conclusion.tex
\section{Conclusion}
\label{sec:conclusion}
In this work, we introduce \textbf{FR3E} (First Return, Entropy-Eliciting Explore), a framework for enhancing exploration in RL for LLMs. 
FR3E identifies high-uncertainty points in reasoning trajectories using token-wise entropy and initiates targeted rollouts to explore alternative solution paths. 
This process constructs intermediate feedback to guide the model, while an adaptive advantage mechanism ensures stable policy updates. 
Results show that FR3E leads to more stable training dynamics, produces longer and more coherent reasoning chains, and increases the yield of correct solutions. 
Overall, FR3E provides an effective and structured approach to improving reasoning in LLMs through more reliable exploration.

\section*{Acknowledgment}
The authors would like to express gratitude to Yingru Li for his insightful review and valuable suggestions for revision of this paper.

%% file: sections/appendix.tex
\section{Data Setting Claim}
\label{app:data_setting}

In our preliminary experiments, we explored alternative data settings. Specifically, we conducted a comparative study on the \textbf{Qwen2.5 7B math base} model. The results indicated that models trained using data from the DAPO dataset yielded inferior performance compared to our current data configuration, which combines the \textbf{DeepScaler} and \textbf{SimpleRL} datasets. 

Consequently, we adopted the current setting for our main experiments. For the sake of experimental consistency and inertia, this data setting was maintained across all subsequent model training processes. However, it is important to note that this conclusion is based on our specific observations with the Qwen2.5 7B math base model. The findings may not be directly transferable or hold true for other base models, such as the \textbf{Qwen2.5 32B base}, which might exhibit different behaviors and data requirements.

%% file: paper.bbl
\begin{thebibliography}{40}
\providecommand{\natexlab}[1]{#1}
\providecommand{\url}[1]{\texttt{#1}}
\expandafter\ifx\csname urlstyle\endcsname\relax
  \providecommand{\doi}[1]{doi: #1}\else
  \providecommand{\doi}{doi: \begingroup \urlstyle{rm}\Url}\fi

\bibitem[Ahmadian et~al.(2024)Ahmadian, Cremer, Gallé, Fadaee, Kreutzer, Pietquin, Üstün, and Hooker]{ahmadian2024basicsrevisitingreinforcestyle}
Arash Ahmadian, Chris Cremer, Matthias Gallé, Marzieh Fadaee, Julia Kreutzer, Olivier Pietquin, Ahmet Üstün, and Sara Hooker.
\newblock Back to basics: Revisiting reinforce style optimization for learning from human feedback in llms, 2024.
\newblock URL \url{https://arxiv.org/abs/2402.14740}.

\bibitem[Auer(2002)]{auer2002using}
Peter Auer.
\newblock Using confidence bounds for exploitation-exploration trade-offs.
\newblock \emph{Journal of Machine Learning Research}, 3\penalty0 (Nov):\penalty0 397--422, 2002.

\bibitem[Auer et~al.(2002)Auer, Cesa-Bianchi, and Fischer]{auer2002finite}
Peter Auer, Nicolo Cesa-Bianchi, and Paul Fischer.
\newblock Finite-time analysis of the multiarmed bandit problem.
\newblock \emph{Machine learning}, 47:\penalty0 235--256, 2002.

\bibitem[Brown et~al.(2020)Brown, Mann, Ryder, Subbiah, Kaplan, Dhariwal, Neelakantan, Shyam, Sastry, Askell, et~al.]{brown2020language}
Tom Brown, Benjamin Mann, Nick Ryder, Melanie Subbiah, Jared~D Kaplan, Prafulla Dhariwal, Arvind Neelakantan, Pranav Shyam, Girish Sastry, Amanda Askell, et~al.
\newblock Language models are few-shot learners.
\newblock \emph{Advances in neural information processing systems}, 33:\penalty0 1877--1901, 2020.

\bibitem[Chen et~al.(2025)Chen, Xu, Liang, He, Pang, Yu, Song, Liu, Zhou, Zhang, Wang, Tu, Mi, and Yu]{chen2025think23overthinkingo1like}
Xingyu Chen, Jiahao Xu, Tian Liang, Zhiwei He, Jianhui Pang, Dian Yu, Linfeng Song, Qiuzhi Liu, Mengfei Zhou, Zhuosheng Zhang, Rui Wang, Zhaopeng Tu, Haitao Mi, and Dong Yu.
\newblock Do not think that much for 2+3=? on the overthinking of o1-like llms, 2025.
\newblock URL \url{https://arxiv.org/abs/2412.21187}.

\bibitem[Cui et~al.(2025{\natexlab{a}})Cui, Yuan, Wang, Wang, Li, He, Fan, Yu, Xu, Chen, Yuan, Chen, Zhang, Lv, Wang, Yao, Han, Peng, Cheng, Liu, Sun, Zhou, and Ding]{cui2025processreinforcementimplicitrewards}
Ganqu Cui, Lifan Yuan, Zefan Wang, Hanbin Wang, Wendi Li, Bingxiang He, Yuchen Fan, Tianyu Yu, Qixin Xu, Weize Chen, Jiarui Yuan, Huayu Chen, Kaiyan Zhang, Xingtai Lv, Shuo Wang, Yuan Yao, Xu~Han, Hao Peng, Yu~Cheng, Zhiyuan Liu, Maosong Sun, Bowen Zhou, and Ning Ding.
\newblock Process reinforcement through implicit rewards, 2025{\natexlab{a}}.
\newblock URL \url{https://arxiv.org/abs/2502.01456}.

\bibitem[Cui et~al.(2025{\natexlab{b}})Cui, Zhang, and et~al.]{cui2025prime}
Yutong Cui, Hongyi Zhang, and et~al.
\newblock Prime: Process reinforcement with implicit model estimation for llms.
\newblock \emph{arXiv preprint arXiv:2504.00052}, 2025{\natexlab{b}}.

\bibitem[Dai et~al.(2025)Dai, Yang, and Si]{dai2025sgrpoearlyexitreinforcement}
Muzhi Dai, Chenxu Yang, and Qingyi Si.
\newblock S-grpo: Early exit via reinforcement learning in reasoning models, 2025.
\newblock URL \url{https://arxiv.org/abs/2505.07686}.

\bibitem[Ecoffet et~al.(2019)Ecoffet, Huizinga, Lehman, Stanley, and Clune]{ecoffet2019go}
Adrien Ecoffet, Joost Huizinga, Joel Lehman, Kenneth~O Stanley, and Jeff Clune.
\newblock Go-explore: a new approach for hard-exploration problems.
\newblock \emph{arXiv preprint arXiv:1901.10995}, 2019.

\bibitem[Ecoffet et~al.(2021)Ecoffet, Huizinga, Lehman, Stanley, and Clune]{Ecoffet_2021}
Adrien Ecoffet, Joost Huizinga, Joel Lehman, Kenneth~O. Stanley, and Jeff Clune.
\newblock First return, then explore.
\newblock \emph{Nature}, 590\penalty0 (7847):\penalty0 580–586, February 2021.
\newblock ISSN 1476-4687.
\newblock \doi{10.1038/s41586-020-03157-9}.
\newblock URL \url{http://dx.doi.org/10.1038/s41586-020-03157-9}.

\bibitem[Forootani(2025)]{forootani2025survey}
Ali Forootani.
\newblock A survey on mathematical reasoning and optimization with large language models.
\newblock \emph{arXiv preprint arXiv:2503.17726}, 2025.

\bibitem[Guo et~al.(2025)Guo, Yang, Zhang, Song, Zhang, Xu, Zhu, Ma, Wang, Bi, et~al.]{guo2025deepseek}
Daya Guo, Dejian Yang, Haowei Zhang, Junxiao Song, Ruoyu Zhang, Runxin Xu, Qihao Zhu, Shirong Ma, Peiyi Wang, Xiao Bi, et~al.
\newblock Deepseek-r1: Incentivizing reasoning capability in llms via reinforcement learning.
\newblock \emph{arXiv preprint arXiv:2501.12948}, 2025.

\bibitem[Huan et~al.(2025)Huan, Li, Zheng, Xu, Kim, Du, Poovendran, Neubig, and Yue]{huan2025doesmathreasoningimprove}
Maggie Huan, Yuetai Li, Tuney Zheng, Xiaoyu Xu, Seungone Kim, Minxin Du, Radha Poovendran, Graham Neubig, and Xiang Yue.
\newblock Does math reasoning improve general llm capabilities? understanding transferability of llm reasoning, 2025.
\newblock URL \url{https://arxiv.org/abs/2507.00432}.

\bibitem[Kazemnejad et~al.(2024)Kazemnejad, Aghajohari, Portelance, Sordoni, Reddy, Courville, and Roux]{kazemnejad2024vineppounlockingrlpotential}
Amirhossein Kazemnejad, Milad Aghajohari, Eva Portelance, Alessandro Sordoni, Siva Reddy, Aaron Courville, and Nicolas~Le Roux.
\newblock Vineppo: Unlocking rl potential for llm reasoning through refined credit assignment, 2024.
\newblock URL \url{https://arxiv.org/abs/2410.01679}.

\bibitem[Li et~al.(2025)Li, Xu, Jiang, Ramasubramanian, Niu, Lin, Yue, and Poovendran]{li2025temporalsamplingforgottenreasoning}
Yuetai Li, Zhangchen Xu, Fengqing Jiang, Bhaskar Ramasubramanian, Luyao Niu, Bill~Yuchen Lin, Xiang Yue, and Radha Poovendran.
\newblock Temporal sampling for forgotten reasoning in llms, 2025.
\newblock URL \url{https://arxiv.org/abs/2505.20196}.

\bibitem[Lightman et~al.(2023)Lightman, Kosaraju, Burda, Edwards, Baker, Lee, Leike, Schulman, Sutskever, and Cobbe]{lightman2023letsverifystepstep}
Hunter Lightman, Vineet Kosaraju, Yura Burda, Harri Edwards, Bowen Baker, Teddy Lee, Jan Leike, John Schulman, Ilya Sutskever, and Karl Cobbe.
\newblock Let's verify step by step, 2023.
\newblock URL \url{https://arxiv.org/abs/2305.20050}.

\bibitem[Lu et~al.(2025)Lu, Hu, and Clune]{lu2025intelligentgoexplorestandingshoulders}
Cong Lu, Shengran Hu, and Jeff Clune.
\newblock Intelligent go-explore: Standing on the shoulders of giant foundation models, 2025.
\newblock URL \url{https://arxiv.org/abs/2405.15143}.

\bibitem[Luo et~al.(2024)Luo, Liu, Liu, Phatale, Guo, Lara, Li, Shu, Zhu, Meng, Sun, and Rastogi]{luo2024improvemathematicalreasoninglanguage}
Liangchen Luo, Yinxiao Liu, Rosanne Liu, Samrat Phatale, Meiqi Guo, Harsh Lara, Yunxuan Li, Lei Shu, Yun Zhu, Lei Meng, Jiao Sun, and Abhinav Rastogi.
\newblock Improve mathematical reasoning in language models by automated process supervision, 2024.
\newblock URL \url{https://arxiv.org/abs/2406.06592}.

\bibitem[Luo et~al.(2025)Luo, Tan, Wong, Shi, Tang, Roongta, Cai, Luo, Li, Popa, and Stoica]{deepscaler2025}
Michael Luo, Sijun Tan, Justin Wong, Xiaoxiang Shi, William~Y. Tang, Manan Roongta, Colin Cai, Jeffrey Luo, Li~Erran Li, Raluca~Ada Popa, and Ion Stoica.
\newblock Deepscaler: Surpassing o1-preview with a 1.5b model by scaling rl.
\newblock \url{https://pretty-radio-b75.notion.site/DeepScaleR-Surpassing-O1-Preview-with-a-1-5B-Model-by-Scaling-RL-19681902c1468005bed8ca303013a4e2}, 2025.
\newblock Notion Blog.

\bibitem[MiniMax et~al.(2025)MiniMax, :, Chen, Li, Gong, Jiang, Fei, Yang, Shan, Yu, Wang, Zhu, Xiao, Du, Zhang, Qiao, Zhang, Du, Guo, Chen, Ding, Sun, Li, Jiao, Zhou, Zhang, Ding, Sun, Feng, Cai, Zhu, Sun, Zhuang, Cai, Song, Zhu, Li, Tian, Liu, Xu, Yan, Liu, He, Feng, Yang, Xiao, Han, Wang, Yu, Feng, Li, Zheng, Du, Yang, Zeng, Yu, Tao, Chi, Zhang, Lin, Hu, Di, Gao, Li, Zhao, Ren, Xu, Li, Wang, Tian, Leng, Chen, Chen, Shi, Weng, Guan, Yu, Li, Zhu, Li, Cai, Liang, Cheng, Kong, Li, Chen, Song, Luo, Su, Li, Han, Hou, Lu, Zou, Shen, Gong, Ma, Wang, Shi, Zhong, Duan, Fu, Hu, Gao, Fan, Yang, Li, Hu, Huang, Li, Xu, Mao, Shi, Wenren, Li, Li, Tian, Zhu, Fan, Wu, Xu, Yu, Lyu, Jiang, Gao, Wu, Song, and Sun]{minimax2025minimaxm1scalingtesttimecompute}
MiniMax, :, Aili Chen, Aonian Li, Bangwei Gong, Binyang Jiang, Bo~Fei, Bo~Yang, Boji Shan, Changqing Yu, Chao Wang, Cheng Zhu, Chengjun Xiao, Chengyu Du, Chi Zhang, Chu Qiao, Chunhao Zhang, Chunhui Du, Congchao Guo, Da~Chen, Deming Ding, Dianjun Sun, Dong Li, Enwei Jiao, Haigang Zhou, Haimo Zhang, Han Ding, Haohai Sun, Haoyu Feng, Huaiguang Cai, Haichao Zhu, Jian Sun, Jiaqi Zhuang, Jiaren Cai, Jiayuan Song, Jin Zhu, Jingyang Li, Jinhao Tian, Jinli Liu, Junhao Xu, Junjie Yan, Junteng Liu, Junxian He, Kaiyi Feng, Ke~Yang, Kecheng Xiao, Le~Han, Leyang Wang, Lianfei Yu, Liheng Feng, Lin Li, Lin Zheng, Linge Du, Lingyu Yang, Lunbin Zeng, Minghui Yu, Mingliang Tao, Mingyuan Chi, Mozhi Zhang, Mujie Lin, Nan Hu, Nongyu Di, Peng Gao, Pengfei Li, Pengyu Zhao, Qibing Ren, Qidi Xu, Qile Li, Qin Wang, Rong Tian, Ruitao Leng, Shaoxiang Chen, Shaoyu Chen, Shengmin Shi, Shitong Weng, Shuchang Guan, Shuqi Yu, Sichen Li, Songquan Zhu, Tengfei Li, Tianchi Cai, Tianrun Liang, Weiyu Cheng, Weize Kong, Wenkai Li, Xiancai Chen,
  Xiangjun Song, Xiao Luo, Xiao Su, Xiaobo Li, Xiaodong Han, Xinzhu Hou, Xuan Lu, Xun Zou, Xuyang Shen, Yan Gong, Yan Ma, Yang Wang, Yiqi Shi, Yiran Zhong, Yonghong Duan, Yongxiang Fu, Yongyi Hu, Yu~Gao, Yuanxiang Fan, Yufeng Yang, Yuhao Li, Yulin Hu, Yunan Huang, Yunji Li, Yunzhi Xu, Yuxin Mao, Yuxuan Shi, Yuze Wenren, Zehan Li, Zelin Li, Zhanxu Tian, Zhengmao Zhu, Zhenhua Fan, Zhenzhen Wu, Zhichao Xu, Zhihang Yu, Zhiheng Lyu, Zhuo Jiang, Zibo Gao, Zijia Wu, Zijian Song, and Zijun Sun.
\newblock Minimax-m1: Scaling test-time compute efficiently with lightning attention, 2025.
\newblock URL \url{https://arxiv.org/abs/2506.13585}.

\bibitem[Ouyang et~al.(2022)Ouyang, Wu, Jiang, and et~al.]{ouyang2022training}
Long Ouyang, Jeff Wu, Xu~Jiang, and et~al.
\newblock Training language models to follow instructions with human feedback.
\newblock In \emph{Advances in Neural Information Processing Systems}, 2022.

\bibitem[Pignatelli et~al.(2023)Pignatelli, Ferret, Geist, Mesnard, van Hasselt, Pietquin, and Toni]{pignatelli2023survey}
Eduardo Pignatelli, Johan Ferret, Matthieu Geist, Thomas Mesnard, Hado van Hasselt, Olivier Pietquin, and Laura Toni.
\newblock A survey of temporal credit assignment in deep reinforcement learning.
\newblock \emph{arXiv preprint arXiv:2312.01072}, 2023.

\bibitem[Qwen et~al.(2025)Qwen, Yang, Yang, Zhang, Hui, Zheng, Yu, Li, Liu, Huang, Wei, Lin, Yang, Tu, Zhang, Yang, Yang, Zhou, Lin, Dang, Lu, Bao, Yang, Yu, Li, Xue, Zhang, Zhu, Men, Lin, Li, Tang, Xia, Ren, Ren, Fan, Su, Zhang, Wan, Liu, Cui, Zhang, and Qiu]{qwen2025qwen25technicalreport}
Qwen, An~Yang, Baosong Yang, Beichen Zhang, Binyuan Hui, Bo~Zheng, Bowen Yu, Chengyuan Li, Dayiheng Liu, Fei Huang, Haoran Wei, Huan Lin, Jian Yang, Jianhong Tu, Jianwei Zhang, Jianxin Yang, Jiaxi Yang, Jingren Zhou, Junyang Lin, Kai Dang, Keming Lu, Keqin Bao, Kexin Yang, Le~Yu, Mei Li, Mingfeng Xue, Pei Zhang, Qin Zhu, Rui Men, Runji Lin, Tianhao Li, Tianyi Tang, Tingyu Xia, Xingzhang Ren, Xuancheng Ren, Yang Fan, Yang Su, Yichang Zhang, Yu~Wan, Yuqiong Liu, Zeyu Cui, Zhenru Zhang, and Zihan Qiu.
\newblock Qwen2.5 technical report, 2025.
\newblock URL \url{https://arxiv.org/abs/2412.15115}.

\bibitem[Ranzato et~al.(2016)Ranzato, Chopra, Auli, and Zaremba]{81a0972a0b704d918ec2ce3696c5b871}
Marc{\textquoteright}Aurelio Ranzato, Sumit Chopra, Michael Auli, and Wojciech Zaremba.
\newblock Sequence level training with recurrent neural networks, 2016.
\newblock Publisher Copyright: {\textcopyright} ICLR 2016: San Juan, Puerto Rico. All Rights Reserved.; 4th International Conference on Learning Representations, ICLR 2016 ; Conference date: 02-05-2016 Through 04-05-2016.

\bibitem[Schulman et~al.(2017)Schulman, Wolski, Dhariwal, Radford, and Klimov]{schulman2017proximalpolicyoptimizationalgorithms}
John Schulman, Filip Wolski, Prafulla Dhariwal, Alec Radford, and Oleg Klimov.
\newblock Proximal policy optimization algorithms, 2017.
\newblock URL \url{https://arxiv.org/abs/1707.06347}.

\bibitem[Schulman et~al.(2018)Schulman, Moritz, Levine, Jordan, and Abbeel]{schulman2018highdimensionalcontinuouscontrolusing}
John Schulman, Philipp Moritz, Sergey Levine, Michael Jordan, and Pieter Abbeel.
\newblock High-dimensional continuous control using generalized advantage estimation, 2018.
\newblock URL \url{https://arxiv.org/abs/1506.02438}.

\bibitem[Setlur et~al.(2024)Setlur, Nagpal, Fisch, Geng, Eisenstein, Agarwal, Agarwal, Berant, and Kumar]{setlur2024rewardingprogressscalingautomated}
Amrith Setlur, Chirag Nagpal, Adam Fisch, Xinyang Geng, Jacob Eisenstein, Rishabh Agarwal, Alekh Agarwal, Jonathan Berant, and Aviral Kumar.
\newblock Rewarding progress: Scaling automated process verifiers for llm reasoning, 2024.
\newblock URL \url{https://arxiv.org/abs/2410.08146}.

\bibitem[Shao et~al.(2025)Shao, Li, Xin, Geng, Wang, Oh, Du, Lambert, Min, Krishna, Tsvetkov, Hajishirzi, Koh, and Zettlemoyer]{shao2025spurious}
Rulin Shao, Shuyue~Stella Li, Rui Xin, Scott Geng, Yiping Wang, Sewoong Oh, Simon~Shaolei Du, Nathan Lambert, Sewon Min, Ranjay Krishna, Yulia Tsvetkov, Hannaneh Hajishirzi, Pang~Wei Koh, and Luke Zettlemoyer.
\newblock Spurious rewards: Rethinking training signals in rlvr.
\newblock \url{https://rethink-rlvr.notion.site/Spurious-Rewards-Rethinking-Training-Signals-in-RLVR-1f4df34dac1880948858f95aeb88872f}, 2025.
\newblock Notion Blog.

\bibitem[Shao et~al.(2024)Shao, Wang, Zhu, Xu, Song, Bi, Zhang, Zhang, Li, Wu, and Guo]{shao2024deepseekmathpushinglimitsmathematical}
Zhihong Shao, Peiyi Wang, Qihao Zhu, Runxin Xu, Junxiao Song, Xiao Bi, Haowei Zhang, Mingchuan Zhang, Y.~K. Li, Y.~Wu, and Daya Guo.
\newblock Deepseekmath: Pushing the limits of mathematical reasoning in open language models, 2024.
\newblock URL \url{https://arxiv.org/abs/2402.03300}.

\bibitem[Sheng et~al.(2025)Sheng, Zhang, Ye, Wu, Zhang, Zhang, Peng, Lin, and Wu]{Sheng_2025}
Guangming Sheng, Chi Zhang, Zilingfeng Ye, Xibin Wu, Wang Zhang, Ru~Zhang, Yanghua Peng, Haibin Lin, and Chuan Wu.
\newblock Hybridflow: A flexible and efficient rlhf framework.
\newblock In \emph{Proceedings of the Twentieth European Conference on Computer Systems}, EuroSys ’25, page 1279–1297. ACM, March 2025.
\newblock \doi{10.1145/3689031.3696075}.
\newblock URL \url{http://dx.doi.org/10.1145/3689031.3696075}.

\bibitem[Sutton et~al.(1998)Sutton, Barto, et~al.]{sutton1998reinforcement}
Richard~S Sutton, Andrew~G Barto, et~al.
\newblock \emph{Reinforcement learning: An introduction}, volume~1.
\newblock MIT press Cambridge, 1998.

\bibitem[Sutton et~al.(1999)Sutton, Precup, and Singh]{sutton1999between}
Richard~S Sutton, Doina Precup, and Satinder~P Singh.
\newblock Between mdps and semi-mdps: A framework for temporal abstraction in reinforcement learning.
\newblock \emph{Artificial intelligence}, 112\penalty0 (1--2):\penalty0 181--211, 1999.

\bibitem[Wang et~al.(2025)Wang, Yu, Gao, Zheng, Liu, Lu, Dang, Chen, Yang, Zhang, Liu, Yang, Zhao, Yue, Song, Yu, Huang, and Lin]{wang20258020rulehighentropyminority}
Shenzhi Wang, Le~Yu, Chang Gao, Chujie Zheng, Shixuan Liu, Rui Lu, Kai Dang, Xionghui Chen, Jianxin Yang, Zhenru Zhang, Yuqiong Liu, An~Yang, Andrew Zhao, Yang Yue, Shiji Song, Bowen Yu, Gao Huang, and Junyang Lin.
\newblock Beyond the 80/20 rule: High-entropy minority tokens drive effective reinforcement learning for llm reasoning, 2025.
\newblock URL \url{https://arxiv.org/abs/2506.01939}.

\bibitem[Wei et~al.(2022)Wei, Wang, Schuurmans, Bosma, Xia, Chi, Le, Zhou, et~al.]{wei2022chain}
Jason Wei, Xuezhi Wang, Dale Schuurmans, Maarten Bosma, Fei Xia, Ed~Chi, Quoc~V Le, Denny Zhou, et~al.
\newblock Chain-of-thought prompting elicits reasoning in large language models.
\newblock \emph{Advances in neural information processing systems}, 35:\penalty0 24824--24837, 2022.

\bibitem[Wu et~al.(2024)Wu, Peng, Du, Zheng, Liu, Wu, Ma, Li, Yang, Zhou, et~al.]{wu2024comparative}
Siwei Wu, Zhongyuan Peng, Xinrun Du, Tuney Zheng, Minghao Liu, Jialong Wu, Jiachen Ma, Yizhi Li, Jian Yang, Wangchunshu Zhou, et~al.
\newblock A comparative study on reasoning patterns of openai's o1 model.
\newblock \emph{arXiv preprint arXiv:2410.13639}, 2024.

\bibitem[Yu et~al.(2025)Yu, Zhang, Zhu, Yuan, Zuo, Yue, Dai, Fan, Liu, Liu, Liu, Lin, Lin, Ma, Sheng, Tong, Zhang, Zhang, Zhang, Zhu, Zhu, Chen, Chen, Wang, Yu, Song, Wei, Zhou, Liu, Ma, Zhang, Yan, Qiao, Wu, and Wang]{yu2025dapoopensourcellmreinforcement}
Qiying Yu, Zheng Zhang, Ruofei Zhu, Yufeng Yuan, Xiaochen Zuo, Yu~Yue, Weinan Dai, Tiantian Fan, Gaohong Liu, Lingjun Liu, Xin Liu, Haibin Lin, Zhiqi Lin, Bole Ma, Guangming Sheng, Yuxuan Tong, Chi Zhang, Mofan Zhang, Wang Zhang, Hang Zhu, Jinhua Zhu, Jiaze Chen, Jiangjie Chen, Chengyi Wang, Hongli Yu, Yuxuan Song, Xiangpeng Wei, Hao Zhou, Jingjing Liu, Wei-Ying Ma, Ya-Qin Zhang, Lin Yan, Mu~Qiao, Yonghui Wu, and Mingxuan Wang.
\newblock Dapo: An open-source llm reinforcement learning system at scale, 2025.
\newblock URL \url{https://arxiv.org/abs/2503.14476}.

\bibitem[Yue et~al.(2025)Yue, Yuan, Yu, Zuo, Zhu, Xu, Chen, Wang, Fan, Du, Wei, Yu, Liu, Liu, Liu, Lin, Lin, Ma, Zhang, Zhang, Zhang, Zhu, Zhang, Liu, Wang, Wu, and Yan]{yue2025vapoefficientreliablereinforcement}
Yu~Yue, Yufeng Yuan, Qiying Yu, Xiaochen Zuo, Ruofei Zhu, Wenyuan Xu, Jiaze Chen, Chengyi Wang, TianTian Fan, Zhengyin Du, Xiangpeng Wei, Xiangyu Yu, Gaohong Liu, Juncai Liu, Lingjun Liu, Haibin Lin, Zhiqi Lin, Bole Ma, Chi Zhang, Mofan Zhang, Wang Zhang, Hang Zhu, Ru~Zhang, Xin Liu, Mingxuan Wang, Yonghui Wu, and Lin Yan.
\newblock Vapo: Efficient and reliable reinforcement learning for advanced reasoning tasks, 2025.
\newblock URL \url{https://arxiv.org/abs/2504.05118}.

\bibitem[Zeng et~al.(2025)Zeng, Huang, Liu, Liu, He, Ma, and He]{zeng2025simplerlzooinvestigatingtamingzero}
Weihao Zeng, Yuzhen Huang, Qian Liu, Wei Liu, Keqing He, Zejun Ma, and Junxian He.
\newblock Simplerl-zoo: Investigating and taming zero reinforcement learning for open base models in the wild, 2025.
\newblock URL \url{https://arxiv.org/abs/2503.18892}.

\bibitem[Zhang et~al.(2025)Zhang, Wang, Cheng, Zhuang, Lin, Zhang, Wang, Cui, Wang, Peng, Jiang, Kuang, Yin, Wen, Zhang, Chen, and Yu]{zhang2025srpocrossdomainimplementationlargescale}
Xiaojiang Zhang, Jinghui Wang, Zifei Cheng, Wenhao Zhuang, Zheng Lin, Minglei Zhang, Shaojie Wang, Yinghan Cui, Chao Wang, Junyi Peng, Shimiao Jiang, Shiqi Kuang, Shouyu Yin, Chaohang Wen, Haotian Zhang, Bin Chen, and Bing Yu.
\newblock Srpo: A cross-domain implementation of large-scale reinforcement learning on llm, 2025.
\newblock URL \url{https://arxiv.org/abs/2504.14286}.

\bibitem[Zhou et~al.(2022)Zhou, Sch{\"a}rli, Hou, Wei, Scales, Wang, Schuurmans, Cui, Bousquet, Le, et~al.]{zhou2022least}
Denny Zhou, Nathanael Sch{\"a}rli, Le~Hou, Jason Wei, Nathan Scales, Xuezhi Wang, Dale Schuurmans, Claire Cui, Olivier Bousquet, Quoc Le, et~al.
\newblock Least-to-most prompting enables complex reasoning in large language models.
\newblock \emph{arXiv preprint arXiv:2205.10625}, 2022.

\end{thebibliography}
